\documentclass[]{TEAI}

\usepackage[utf8]{inputenc}
\usepackage[T1]{fontenc}
\usepackage{url}
\usepackage{booktabs}
\usepackage{array}
\usepackage{multirow}
\usepackage{longtable}
\usepackage{amsfonts}
\usepackage{amsmath}
\usepackage{nicefrac}
\usepackage{graphicx}
\usepackage{float}
\usepackage{algorithm}
\usepackage{algpseudocode}
\usepackage{tikz}
\usetikzlibrary{arrows.meta,positioning,shapes.geometric,fit,calc}
\usepackage{pgfplots}
\usepackage{listings}
\pgfplotsset{compat=1.18}

\lstdefinestyle{promptblock}{
  basicstyle=\footnotesize\ttfamily,
  breaklines=true,
  breakatwhitespace=false,
  columns=fullflexible,
  keepspaces=true,
  showstringspaces=false,
  xleftmargin=0pt,
  xrightmargin=0pt,
  aboveskip=0.35\baselineskip,
  belowskip=0.35\baselineskip
}

\newcommand{\method}{\textsc{VeriEvol}}
\newcommand{\agent}{\textsc{HTV-Agent}}
\newcommand{\sftdata}{\textsc{VeriEvol-SFT}}
\newcommand{\rldata}{\textsc{VeriEvol-RL}}

\definecolor{evolpre}{HTML}{8A8F99}    
\definecolor{evolpost}{HTML}{6B46C1}   
\definecolor{evolprebg}{HTML}{F2F3F5}  
\definecolor{evolpostbg}{HTML}{F3EFFB} 
\newtcolorbox{seedbox}{
  enhanced, breakable, boxrule=0.4pt, arc=2pt,
  colframe=evolpre, colback=evolprebg, colbacktitle=evolpre,
  left=5pt, right=5pt, top=3pt, bottom=3pt,
  fonttitle=\bfseries\footnotesize\color{white},
  title={\textsf{Before evolution \,--\, seed prompt}}
}
\newtcolorbox{evolbox}{
  enhanced, breakable, boxrule=0.6pt, arc=2pt,
  colframe=evolpost, colback=evolpostbg, colbacktitle=evolpost,
  left=5pt, right=5pt, top=3pt, bottom=3pt,
  fonttitle=\bfseries\footnotesize\color{white},
  title={\textsf{After evolution \,--\, image-grounded reasoning prompt}}
}
\newcommand{\ansok}[1]{\textcolor{evolpost}{\textbf{#1}}}

\newcommand{\opsep}{\,\textperiodcentered\,}                 
\newcommand{\taxhead}[1]{\textbf{#1}}                        
\newcommand{\taxrow}[1]{\textbf{#1}}                         
\newcommand{\taxsep}{\noalign{\vskip2pt}\arrayrulecolor{black!35}\hline\arrayrulecolor{black}\noalign{\vskip2pt}}

\title{\method: Scaling Multimodal Mathematical Reasoning\\
via Verifiable Evol-Instruct}

\author{
    Haoling Li\textsuperscript{1,2,*},
    Kai Zheng\textsuperscript{2,*,$\dagger$},
    Jie Wu\textsuperscript{1},\\
    Can Xu\textsuperscript{2,$\dagger$},
    Qingfeng Sun\textsuperscript{2},
    Han Hu\textsuperscript{2},
    Yujiu Yang\textsuperscript{1,$\dagger$}
}

\affiliation[1]{\mbox{Tsinghua University}} 
\affiliation[2]{\mbox{Tencent Hunyuan}

\textsuperscript{*}Equal contribution, \textsuperscript{$\dagger$}Corresponding authors\\[-1.7em]}
\correspondence{\textcolor{seedblue}{li-hl23@mails.tsinghua.edu.cn,\{kevinezheng,leocaxu\}@tencent.com,yang.yujiu@sz.tsinghua.edu.cn}}
\checkdata[Website]{\url{https://robertmarton.github.io/verievol}}

\abstract{Scaling reinforcement learning for visual mathematical reasoning requires more than generating harder questions: as data volume grows, the reward labels themselves must remain reliable. Yet existing data pipelines scale supervision while trusting the labeller, and policy-side methods assume the underlying answers are already correct. We instead treat scaling as a verifiable data-construction problem and decouple two axes before any policy update: prompt difficulty, expanded by route-specific evolution operators, and answer reliability, enforced by offline hypothesis--test falsification. We instantiate this as \method, an iterative framework with two extensible components: a type-aware evolution module that rewrites low-difficulty image-question seeds into harder, image-grounded prompts; and \agent, a verifier that accepts an answer only after multi-source counter-evidence has failed to refute it. The resulting verified data scales in volume, extends by adding evolution routes or verifier channels, and plugs directly into existing GRPO-style RL recipes. On a five-benchmark visual-math suite, scaling evolved SFT data from 10K to 250K samples raises the mean accuracy from 35.42 to 54.73; then, with backbone, SFT initialization, and GRPO recipe held fixed, \method\ adds a cumulative $+3.88$ over an un-evolved RL baseline, of which $+1.82$ comes from evolved prompts and $+2.06$ from the \agent\ verifier. We release the prompts, data, models, code, and the full verifier trace of every sample, so that downstream work can scale and audit the pipeline rather than only inspect its outputs.}

\begin{document}

\maketitle

\begingroup
\centering
\vspace{-0.5em}
\includegraphics[width=0.85\linewidth]{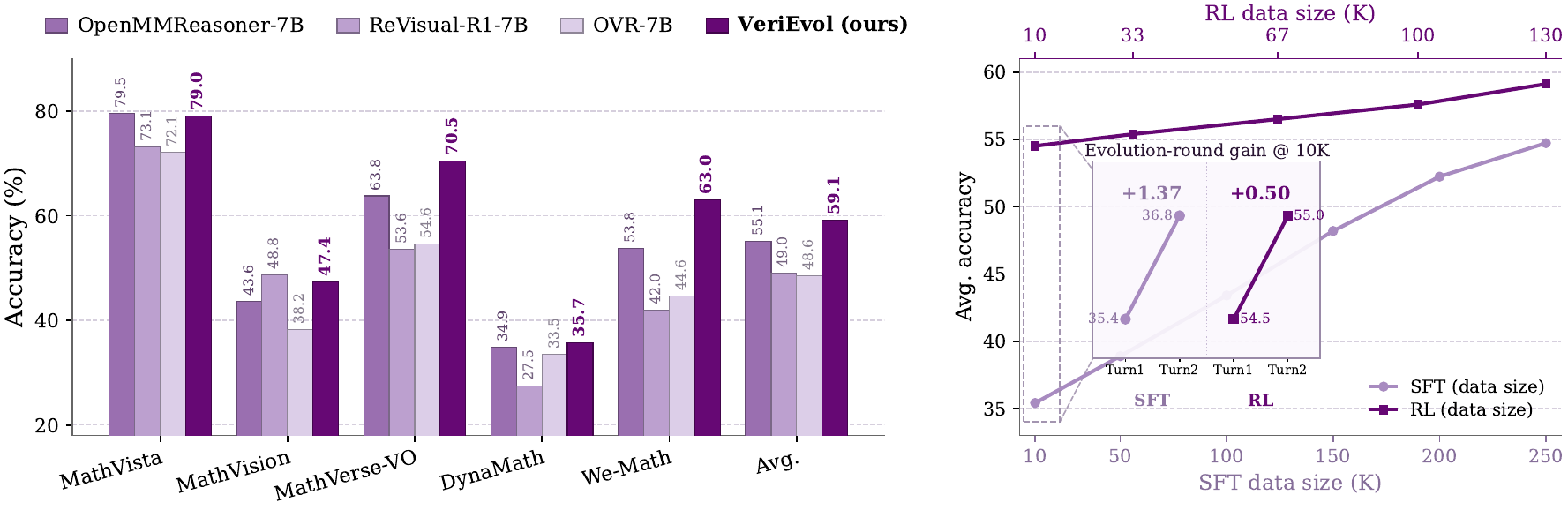}
\vspace{-0.7em}
\captionof{figure}{\textbf{External comparison (left) and data-volume scaling (right).} \method\ leads three of five benchmarks and the mean against the strongest external 7B baselines; evolved SFT data ($10\!\to\!250$K) and verified RL data ($10\!\to\!130$K) both scale monotonically. Details in \S\ref{sec:exp_main}--\S\ref{sec:exp_ablation}.}
\label{fig:firstpage_scaling}
\vspace{-0.4em}
\par
\endgroup

\section{Introduction}

Reasoning-centric progress in large language models, exemplified by RL-based systems like DeepSeek-R1 \citep{deepseekr1}, has made data quality and reward quality first-order training bottlenecks. The issue is sharper for visual reasoning, where a model must ground variables in diagrams, read values from charts, align symbols with visual structures, and execute multi-step mathematical inference. Existing benchmarks \citep{mathvista,mathverse,mmmu,mathvision,mvmath} show that current multimodal large language models (MLLMs) still fall short on these compositional skills.

Scaling this training regime is not only a matter of collecting more prompts. Many multimodal question-answer pairs are image-grounded but of low difficulty, emphasizing recognition or local extraction over multi-step reasoning; generic synthetic pipelines can increase corpus size while producing questions that are merely longer, weakly visual, or answerable from text priors. The bottleneck becomes more severe at the RL stage: as the number of rollouts and training samples grows, any unverified label is repeatedly converted into reward signal, so label noise is no longer a static annotation error but a training objective that can systematically reinforce incorrect policies.

Two complementary lines of work address parts of this problem. Data-pipeline work \citep{openmmreasoner,mmr1,mmevol} scales supervision while trusting the labeller, leaving the answer set unverified; policy-side work \citep{mmeureka,visionr1,vlrethinker,papo} reshapes the supervision distribution or modifies GRPO updates while assuming the underlying answers are correct. Both families leave the answer set's reliability to be fixed \emph{inside} the policy/training loop. We instead make it a property of the data: prompt difficulty scales through route-specific evolution, and---what neither family does explicitly---every answer must survive offline, refutation-seeking hypothesis--test falsification \emph{before} it ever becomes a reward, with this verification decoupled from the policy optimizer. Evolving prompts is shared with prior data-pipeline work; the offline answer-falsification gate is the part that is new. We instantiate this design as \method, an iterative framework with two extensible components: type-aware prompt evolution---which routes each seed by problem family (e.g.\ geometry, charts, OCR) to family-specific operators (\S\ref{sec:evolution})---and \agent, a hypothesis--test verifier that accepts an answer only after multi-source counter-evidence has failed to refute it (\S\ref{sec:agents}). Because the output is a standard verified prompt--answer--reward tuple, the resulting data can be expanded without changing existing GRPO-style RL recipes.

Our contributions are threefold. \textbf{(i)} A scalable multimodal evol-instruct framework that transforms low-difficulty image-question seeds into harder, image-grounded prompts usable for both SFT and RL; in our experiments, increasing SFT data from 10K to 250K raises the mean by $+19.31$; at the RL stage, scaling verified data from 10K to 130K adds a smooth $+4.60$ along the \emph{data-volume} axis, while at a fixed 130K budget the \emph{design} itself contributes $+3.88$ over an un-evolved RL baseline (decomposed as $+1.82$ from evolved prompts and $+2.06$ from the verifier). \textbf{(ii)} \agent, a modular hypothesis--test agent-as-a-judge that screens each answer through four stages---independent solver hypotheses; refutation-seeking verification across counter-evidence, programmatic, and visual channels; conflict resolution by a decider; and a deterministic acceptance gate---so that new verifier channels or answer contracts can be added without re-engineering the policy optimizer. \textbf{(iii)} A traceable open release for extensible data construction: in addition to checkpoints and prompts, every training sample is released with its full \emph{verifier trace} (solver hypotheses, counter-evidence reports, tool outputs, decider rationale, per-component gate outcomes), so that downstream work can audit, extend, and rescale construction decisions rather than only inspect final outputs.

\section{Related Work}

\paragraph{Benchmarks and training data for multimodal reasoning.}
Canonical benchmarks such as MathVista, MathVerse, MMMU, and MATH-Vision \citep{mathvista,mathverse,mmmu,mathvision} serve as the principal evaluation anchors for multimodal reasoning. They are complemented by OlympiadBench \citep{olympiadbench}, DynaMath \citep{dynamath}, We-Math \citep{wemath}, and multi-visual math \citep{mvmath}, which probe olympiad-level difficulty, input perturbation, knowledge structure, and multi-image composition, respectively; multilingual coverage is further probed by M3Kang \citep{m3kang}. On the training side, Qwen2.5-VL \citep{qwen25vl}, MAmmoTH-VL \citep{mammothvl}, VisualWebInstruct \citep{visualwebinstruct}, and MathCoder-VL \citep{mathcodervl} progressively strengthen backbones and SFT corpora via cross-modal chain-of-thought, web mining, and code-aided rendering. These efforts remain at SFT scale, however, and do not provide answer verification of RL-grade reliability---i.e., reliable enough to be used as a reward signal, where a wrong label is reinforced across many rollouts rather than seen once. Honey-Data-15M \citep{honeydata15m} and the multimodal reasoning corpus of \citet{mmfinereason} are adopted in our experimental setting as disjoint seed pools for SFT and RL, respectively. Both supply large-scale, image-grounded STEM questions, but their answers are predominantly of low difficulty and of insufficient reliability for direct RL supervision.

\paragraph{Prompt evolution for reasoning.}
Prompt-evolution methods address the difficulty gap. Evol-Instruct \citep{wizardlm} and MMEvol \citep{mmevol} systematically rewrite seed instructions into more complex variants, and subsequent data-centric efforts extend this paradigm to multimodal training. Representative examples include MathBook-7B \citep{mathbook}, which constructs knowledge-tree-grounded curricula, and OpenMMReasoner \citep{openmmreasoner}, which combines 874K evolution-style SFT samples with GSPO-based RL. By contrast, \method\ jointly performs instruction evolution and explicit answer-level falsification, and produces both SFT and RL data within a single, fully traceable construction pipeline.

\paragraph{RL training: reward design, policy optimization, and answer verification.}
The reliability of the reward signal is a central concern across recent reasoning systems, ranging from text-domain RL such as DeepSeek-R1 and Kimi k1.5 \citep{deepseekr1,kimi15} to multimodal judges \citep{mrjudge,judgeanything}. We group prior efforts into two families. The first family makes the signal more reliable \emph{from inside the RL loop}, along three complementary directions: \textbf{(a) reward design and supervision shape}---visual reinforcement fine-tuning \citep{visualrft} and rule-based rewards on rendered geometry and chart tasks \citep{mmeureka,r1v} target answer-extraction correctness, while cold-start reasoning traces \citep{visionr1,r1onevision,avar} and self-reflection chains \citep{vlrethinker} reshape the supervision distribution before or alongside RL; a parallel thread shapes the \emph{visual} signal itself, via perception-anchored attention supervision \citep{ease,bips}, image-interactive rollouts that ``think with images'' \citep{deepeyes}, and trajectory-aware intrinsic reasoning supervision for self-evolving models \citep{ireasoner}; \textbf{(b) policy optimization} for vision-language models---Skywork-R1V2 \citep{skyworkr1v2}, OVR \citep{ovr}, DUPL \citep{dupl}, PerL-VL \citep{perlvl}, VGPO \citep{vgpo}, and Infi-MMR \citep{infimmr} modify GRPO-based updates through reward shaping, advantage normalization, length penalties, or curriculum scheduling; and \textbf{(c) data-side sampling}---MMR1 \citep{mmr1} reweights rollouts by variance to stabilize GRPO, and NoisyRollout \citep{noisyrollout} augments rollout trajectories with controlled visual perturbations. The second family instead equips models with \emph{explicit verification}: at inference time, Self-Refine \citep{selfrefine}, Reflexion \citep{reflexion}, SelfCheckGPT \citep{selfcheckgpt}, and CRITIC \citep{critic} revise drafts via self-feedback or external tools; at training time, PRM800K \citep{prm800k}, Math-Shepherd \citep{mathshepherd}, and ReST-EM \citep{restem} construct reward-grade supervision via human step-level labels, automatic per-step verification, or self-generated-then-filtered EM loops.

A common assumption across both families is that the underlying answer set is either correct or correctable from inside the policy/training loop; \method\ is orthogonal: it performs answer verification at the data-construction stage (before the policy update), targeting visual-reasoning failure modes that step-level text verifiers cannot diagnose, specifically the visual grounding failures (VG) and text-only shortcuts (TS) we taxonomize in Section~\ref{sec:exp_error}. \method\ shares with the inference-time refinement work the principle that a candidate answer is accepted only after attempts to refute it with counter-evidence have failed, and with the training-time text-domain work the practice of performing verification offline; the resulting \rldata\ is directly compatible with existing policy-optimization and sampling techniques without modification.

\section{\method: Evolving Prompts, Falsifying Answers}
\label{sec:method}

\begin{figure*}[t]
\centering
\includegraphics[width=\linewidth]{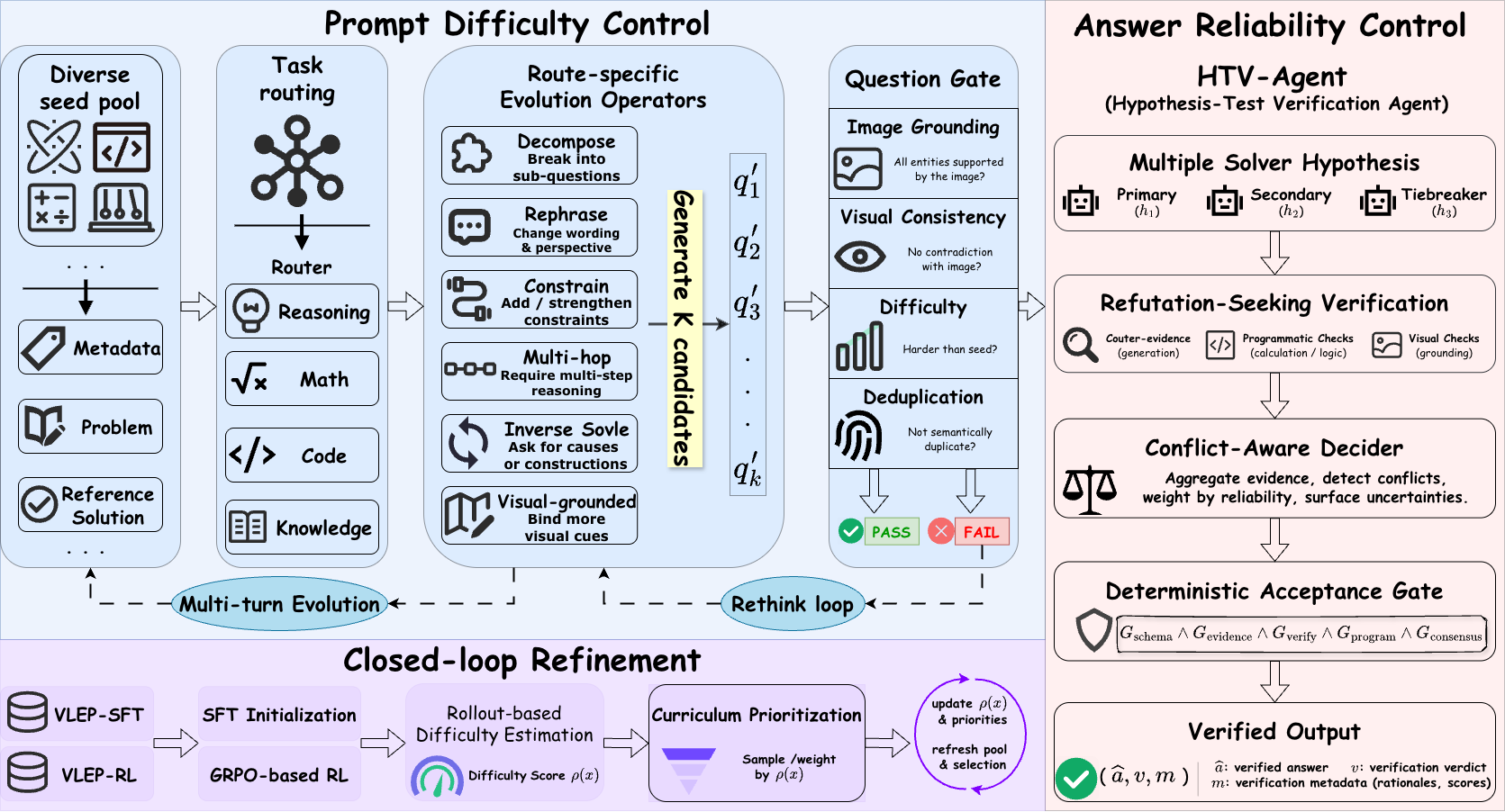}
\caption{\textbf{\method\ framework.} Two core components turn low-difficulty image-question seeds into verified training samples inside a closed loop. \textbf{(Top-left, blue) Prompt Difficulty Control} evolves seeds and filters candidates through a question gate. \textbf{(Right, orange) Answer Reliability Control (\agent)} falsifies candidate answers and admits them through a deterministic gate. \textbf{(Bottom-left, purple) Closed-loop Refinement} materializes accepted samples into \sftdata/\rldata\ and recycles them into the seed pool. Stages detailed in \S\ref{sec:evolution}, \S\ref{sec:agents}, and \S\ref{sec:training_signal}.}
\label{fig:verievol-framework}
\end{figure*}

\method\ takes low-difficulty image-question seeds and produces training samples whose prompts are harder, whose answers have been independently verified, and whose construction decisions are fully traceable. The framework is organized around a scaling principle: separate the components that should grow along different axes. Prompt difficulty scales by adding routes and evolution operators; answer reliability scales by adding independent refutation channels and deterministic gates; training scale grows by materializing the accepted outputs as standard prompt--answer--reward tuples for existing SFT and GRPO recipes. This separation avoids a single end-to-end generator whose difficulty, verification, and reward interfaces fail together, and makes each component separately auditable as the data pool expands. Figure~\ref{fig:verievol-framework} shows the two core components, the closed-loop deployment, and their interfaces. \S\ref{sec:problem} defines the data objects; \S\ref{sec:evolution} describes how route-specific operators rewrite seed prompts; \S\ref{sec:agents} introduces \agent, the hypothesis--test verifier, and its deterministic acceptance gate; and \S\ref{sec:training_signal} specifies how verified samples are materialized into SFT and RL training signal.

\subsection{Problem Formulation}
\label{sec:problem}

A seed sample is denoted $x{=}(I,q,a)$, where $I$ is an image, $q$ is a seed question, and $a$ is an optional seed answer. Our objective is to construct a training sample $\tilde{x}{=}(I,q',\hat{a},m)$, where $q'$ is a more challenging yet still image-grounded prompt, $\hat{a}$ is a verified answer, and $m$ records routing tags, evidence, tool traces, verifier outputs, acceptance decisions, reward-checker specifications, and rollout statistics. The framework returns two data products, $\mathcal{D}_{\mathrm{SFT}}{=}\{(I,q',r,\hat{a},e)\}$ and $\mathcal{D}_{\mathrm{RL}}{=}\{(I,q',\hat{a},v,m)\}$, where $r$ is a reasoning trace, $e$ is supporting evidence, and $v$ is the verification verdict supporting $\hat{a}$. A sample can be useful for SFT without being admitted to \rldata; the latter requires stronger answer normalization and deterministic reward-checker guarantees. The full pseudocode of the iterative construction algorithm is provided in Appendix~\ref{app:algorithm}.

Two design axes are varied independently. \textbf{Prompt difficulty} is controlled solely by route-specific operators $\mathcal{O}_z$ acting on the question, $q\!\to\!q'$, while leaving the answer channel untouched; \textbf{answer reliability} is controlled solely by the verifier $v$ acting on the candidate answer $\hat{a}$, independently of how $q'$ was produced. This separation is what lets the two axes be scaled and ablated on their own: empirically, their RL-stage contributions are near-additive (Section~\ref{sec:exp_main}), which is the evidence that they act on orthogonal failure modes rather than a shared one.

\subsection{Type-Aware Prompt Evolution}
\label{sec:evolution}

\paragraph{Routing.} A single evolution template applied uniformly to all images either over-specifies (e.g., a geometry-style rewrite applied to a chart) or under-specifies (e.g., a generic "make harder" applied to OCR). To match the operator to the image content, we route every sample into a structured tag tuple $z=(t,y,u,c)$, where $t$ is a coarse problem family, $y$ is the answer type, $u$ is a recommended tool set, and $c$ is a compact visual-evidence sketch. The framework figure shows coarse high-level routing categories for readability; the complete visual-route-conditioned taxonomy used to define the operator space is given in Appendix~\ref{app:operators}. The router is implemented as a schema-constrained classifier rather than unconstrained captioning; low-confidence routes fall back to a conservative generic math route or are regenerated with an auxiliary image caption.

\paragraph{Operators.} The operator is the unit that produces the harder prompt; we make it route-specific so that the reasoning-skill amplification, the answer format, and the visual-evidence requirement are all tied to the routed tags rather than chosen at generation time. For each route $z$, \method\ samples an evolution operator from a route-specific set $\mathcal{O}_z$. Each operator is a constrained prompt template specifying (i) the reasoning skill to amplify, (ii) the visual evidence that must remain necessary, (iii) the desired answer format, and (iv) generator failure modes to avoid. Given $K$ candidates, we score each by image dependence, mathematical lift over $q$, answerability, verifiability, and a penalty for prompts that turn out to be solvable from text alone; the top candidates are passed to question-level filtering subject to diversity constraints. Figure~\ref{fig:evol_cases} in Appendix~\ref{app:evol_cases} shows representative before/after pairs, illustrating how a single-step recognition seed is rewritten into a harder, image-grounded reasoning prompt with a short, verifiable answer.

\paragraph{Question-level gate.} A generator that scores its own outputs will systematically reward whatever style it produces, so we filter generated prompts with a gate that is independent of the generator: $\phi_q(I,q,q')=\mathbf{1}[C_{\mathrm{img}} \land C_{\mathrm{cons}} \land C_{\mathrm{vis}} \land C_{\mathrm{hard}} \land C_{\mathrm{dedup}}]$. The five checks require that $q'$ is grounded in visible image evidence, is visually consistent with the image, fails a text-only probe so that the image remains necessary, introduces a non-trivial difficulty lift over $q$, and is not a near-duplicate of existing accepted prompts. Schema fields, full operator taxonomy, per-check thresholds, and scoring weights are given in Appendix~\ref{app:operators}.

\subsection{\agent: Hypothesis--Test Answer Verification}
\label{sec:agents}

After a prompt passes $\phi_q$, \agent\ takes the prompt through four stages: hypothesis generation by independent solvers, refutation-seeking verification with counter-evidence, programmatic, and visual channels, conflict resolution by a decider, and a deterministic acceptance gate. Two design choices are central throughout. \emph{Multiple independent solvers} expose disagreement that a single-solver pipeline cannot, supplying the candidates a verifier needs to compare. \emph{Verifiers asked to refute rather than confirm} probe failure modes a solver has structural reasons to overlook (visual grounding failures, text-only shortcuts, arithmetic slips, in the taxonomy of Section~\ref{sec:exp_error}). The remaining components, the decider and the gate, consume these signals; they do not generate new evidence. Each first answer is therefore treated as a falsifiable hypothesis, never as ground truth. The four stages and the deterministic gate are depicted as the orange ``Answer Reliability Control'' panel of Figure~\ref{fig:verievol-framework}; the loop in the framework is the prompt-rewrite loop used before answer verification when a generated question fails $\phi_q$.

\paragraph{Solvers.}
We run three solver branches at sampling temperatures $\{0.2,0.6,0.15\}$: a low-temperature primary solver ($T{=}0.2$), a higher-temperature secondary solver ($T{=}0.6$) to encourage independent reasoning trajectories, and a low-temperature tiebreaker ($T{=}0.15$) that is invoked only when the first two disagree. Each branch returns a tuple $h_b=(a_b,r_b,e_b,\kappa_b)$, where $a_b$ is the candidate answer, $r_b$ the reasoning trace, $e_b$ the claimed evidence, and $\kappa_b$ a self-reported confidence after normalization.

\paragraph{Verifiers: counter-evidence with programmatic and visual checks.}
For each candidate cluster, a verifier is prompted to actively look for counter-evidence rather than to confirm the answer: it performs blind elimination on multiple-choice prompts, and back-substitutes the candidate answer into the prompt constraints to search for contradictions on numerical or symbolic prompts. Two complementary evidence sources back the verifier. A code-checking module translates the relevant arithmetic, algebraic, or combinatorial constraints into Python assertions executed under a restricted interpreter and an AST-safe arithmetic evaluator. A visual-evidence module checks that the cited quantities and symbols are grounded in $I$ via OCR, local crops, and pixel-level structural probes that return measurements such as bounding-box statistics or row/column intensity profiles. Programmatic checks and visual checks are kept as distinct evidence channels because each catches errors the other misses; the evidence-channel taxonomy is summarized in Figure~\ref{fig:verievol-framework}, and the concrete tool list is detailed in Appendix~\ref{app:operators}.

\paragraph{Conflict-aware decider.}
The decider receives the solver hypotheses together with verifier reports, programmatic results, and visual evidence, and decides whether each objection is logically valid and evidence-grounded. It does not assume that the verifier is more reliable than the solver: when verification refutes the solver with sufficient evidence, the answer is revised; when the verifier is inconsistent or unsupported, the solver answer is retained; when neither side is reliable, the sample is rejected rather than forced into the dataset.

Concretely, the decider is a constrained LLM call (no tool use, no code execution) whose prompt asks the model to (i) summarize the verifier's conclusion in one or two sentences, (ii) assess the \emph{quality} of the verification (logical soundness, self-contradictions, arithmetic errors, unsupported claims) rather than only its conclusion, (iii) commit to a final answer under the three rules above and emit it inside a \texttt{<final\_answer>} tag, (iv) emit an integer confidence in $[0,100]$, and (v) optionally emit \texttt{<require\_rethink>true</require\_rethink>} as an abstention flag. Such low-confidence conflicts are treated as non-admissions by the downstream gate. The verbatim decider prompt is reproduced in Appendix~\ref{app:decider-prompt}.

\paragraph{Deterministic acceptance gate.}
Candidate answers are first normalized into a canonical form (resolving percentage/decimal equivalence, option letter vs.\ option text, simple symbolic equivalents, and unit-bearing numbers) and then passed through a deterministic answer gate
\begin{equation}
g(\hat{a},m)=\mathbf{1}[G_{\mathrm{schema}}\land G_{\mathrm{evidence}}\land G_{\mathrm{verify}}\land G_{\mathrm{program}}\land G_{\mathrm{consensus}}],
\end{equation}
whose components check, respectively, output format ($G_{\mathrm{schema}}$), visual support ($G_{\mathrm{evidence}}$), verifier approval ($G_{\mathrm{verify}}$), executable assertions ($G_{\mathrm{program}}$), and solver-cluster agreement ($G_{\mathrm{consensus}}$). The gate is conjunctive by design: any single failing channel rejects the sample, so admission requires agreement from all evidence sources rather than majority voting.

\subsection{From Verified Answers to Training Signal}
\label{sec:training_signal}

\paragraph{Materialization for SFT and RL.}
Accepted samples are materialized differently depending on how they will be used. For \sftdata, we store $(I,q',r,\hat{a},e)$: the supervised target is a complete response, so the reasoning trace $r$ and supporting evidence $e$ are retained as imitation signal together with the verified final answer. For \rldata, we store $(I,q',\hat{a},v,m)$: the policy must generate its own reasoning during rollout, so the supervised trace is removed, $v$ records the verification verdict supporting $\hat{a}$, and $m$ carries the canonical answer form and reward-checker specification. This makes the admission bar for \rldata stricter than for \sftdata: an SFT example may be useful when its verified rationale and evidence are high quality, whereas an RL example must additionally admit a deterministic checker such as exact match, numeric tolerance, option matching, symbolic equivalence, or executable assertions.

\paragraph{Reward and rollout-based curriculum.}
Let $\nu_m$ denote the deterministic reward checker reconstructed from the metadata $m$. The simplest RL reward is $R(y\mid I,q')=\mathbf{1}[\nu_m(y)=\hat{a}]-\beta\mathbf{1}[\mathrm{format\_error}(y)]$; metadata also supports richer features (verifier confidence, program-assertion status, cluster size). Finally, after training a student checkpoint we estimate $\rho(x)=\frac{1}{N}|\{i \in [N] : \nu_m(y_i)=\hat{a}\}|$ over $N$ rollouts; examples with intermediate $\rho$ are prioritized for subsequent RL, while $\rho{=}1$ items are dropped as trivially solvable and $\rho{=}0$ items as intractable or noisy. The updated $\rho(x)$ values are fed back into seed-pool selection, so later evolution rounds focus on prompts that are neither already solved nor too noisy to verify.

\section{Experiments}
\label{sec:exp}

The experiments evaluate both the effectiveness of \method\ and the scaling properties that motivate the framework:
(i) at fixed backbone and optimizer, evolved prompts and verified answers should improve SFT and RL supervision over seed or unverified data (\S\ref{sec:exp_main});
(ii) supervision should remain useful as SFT data grows from 10K to 250K samples and verified RL data grows from 10K to 130K samples (\S\ref{sec:exp_ablation});
(iii) the verifier should behave as a modular reliability layer, with complementary evidence channels and residual errors shifting away from visual-grounding failures (\S\ref{sec:exp_ablation}, \S\ref{sec:exp_error}).

\subsection{Experimental Setup}
\label{sec:exp_setup}

\paragraph{Models, data, and training.}
The student is Qwen2.5-VL-7B-Instruct \citep{qwen25vl} throughout; the proprietary models used inside \method\ (Seed-2.0-Pro for the SFT branch, Gemini-3-flash for the RL branch) act only as data-construction teachers and are never deployed. All RL runs start from a shared checkpoint \textbf{VeriEvol-SFT-Init}, obtained by SFT on the 250K-sample \sftdata\ (STEM seeds from Honey-Data-15M \citep{honeydata15m}); the SFT data-volume sweep trains the same backbone on smaller \sftdata\ subsets before any RL. RL is then GRPO on top of this init with a binary correctness reward (Appendix~\ref{app:compute} for the full recipe), using the 130K-sample \rldata\ pool whose seeds come from the multimodal reasoning corpus of \citet{mmfinereason} and are restricted to programmatically verifiable answers.

\paragraph{Three RL pipelines and bound argument.}
We compare three pipelines that differ \emph{only} in the RL data: \textbf{RL-Origin} (un-evolved seed prompts), \textbf{RL-Evol} (\method-evolved prompts, model answers as supervision), and \textbf{RL-Evol+Verifier} (the full \method, with \agent-verified answers). All three share VeriEvol-SFT-Init and the GRPO recipe, so the RL-stage deltas directly isolate the \emph{marginal} effect of evolution and verification when applied at the RL stage; the SFT-stage contribution of evolution is reported separately by the SFT rows ($+2.93$ over Seed-only SFT) and is not part of the $+3.88$ RL-stage delta. The remaining confound is an interaction between the evolved init and RL-stage data construction (an evolved init may be more or less responsive to evolved RL data than a seed init); we discuss this residual in Section~\ref{sec:limits} and do not claim a fully independent decomposition.

\paragraph{Baselines, benchmarks, and protocol.}
We compare against recent 7B-size visual reasoning baselines: OpenMMReasoner \citep{openmmreasoner}, OVR \citep{ovr}, ReVisual-R1 \citep{revisualr1}, MMR1 \citep{mmr1}, WeThink \citep{wethink}, VLAA-Thinker \citep{vlaathinker}, VL-Rethinker \citep{vlrethinker}, ThinkLite-VL \citep{thinklitevl}, MM-Eureka \citep{mmeureka}, OpenVLThinker \citep{openvlthinker}, TRON \citep{tron}, Vision-R1 \citep{visionr1}, PaLMR \citep{palmr}, NoisyRollout \citep{noisyrollout}, V-Zero \citep{vzero}, and the Qwen2.5-VL-7B-Instruct \citep{qwen25vl} and InternVL3-8B \citep{internvl3} backbones. Their scores are taken from the original papers or benchmark-complete comparison tables (not re-run), with one exception: OpenMMReasoner-7B's We-Math-Strict score, which we reproduce under our harness because its paper reports only the loose metric. We evaluate on five held-out benchmarks (MathVista\_MINI \citep{mathvista}, MathVision\_MINI, MathVerse Vision-Only \citep{mathverse}, DynaMath-Worst \citep{dynamath}, and We-Math-Strict \citep{wemath}) using each benchmark's official metric under VLMEvalKit ($T{=}0.6$, mean over three runs, cross-run $\sigma{<}0.5$ on every benchmark). For brevity we use the short names MathVista, MathVision, MathVerse-VO, DynaMath, and We-Math in prose, while tables carry the explicit split/metric suffixes (\_MINI, Vision-Only/VO, -Worst, -Strict); these refer to the same evaluation throughout. We single out MathVerse-VO because it most directly tests image reliance. Protocol markers ($\dagger$) for external entries and the within-row $\mathrm{Avg.}$ definition are documented in the caption of Table~\ref{tab:mainresults}.

\subsection{Main Results}
\label{sec:exp_main}

\begin{table}[t]
\centering
\scriptsize
\setlength{\tabcolsep}{4pt}
\renewcommand{\arraystretch}{1.10}
\resizebox{\linewidth}{!}{%
\begin{tabular}{lcccccc}
\toprule
\textbf{Method} & \shortstack{MathVista\\Mini\,$\uparrow$} & \shortstack{MathVision\\Mini\,$\uparrow$} & \shortstack{MathVerse\\VO\,$\uparrow$} & \shortstack{DynaMath\\Worst\,$\uparrow$} & \shortstack{We-Math\\Strict\,$\uparrow$} & $\mathrm{Avg.}\,\uparrow$ \\
\midrule
\multicolumn{7}{c}{\emph{External 7B-size baselines.}} \\[1pt]
OpenMMReasoner-7B \citep{openmmreasoner} & \textbf{79.50} & 43.60 & 63.80 & 34.90 & 53.81 & 55.12 \\
ReVisual-R1-7B \citep{revisualr1} & 73.10 & \textbf{48.80} & 53.60 & 27.50 & 42.00 & 49.00 \\
OVR-7B \citep{ovr} & 72.10 & 38.20 & 54.60 & 33.50 & 44.60 & 48.60 \\
MMR1-Math-v0 \citep{mmr1} & 72.00 & 29.00 & 55.40 & 27.90 & 31.90 & 43.24 \\
WeThink-7B \citep{wethink} & 70.90 & 27.20 & 44.70 & 24.40 & 48.00 & 43.04 \\
VLAA-Thinker-7B \citep{vlaathinker} & 68.00 & 26.40 & 48.20 & 22.40 & 41.50 & 41.30 \\
VL-Rethinker-7B \citep{vlrethinker} & 73.70 & 28.40 & 46.40 & 17.80 & 36.30 & 40.52 \\
ThinkLite-VL-7B \citep{thinklitevl} & 71.60 & 24.60 & 42.90 & 16.50 & 41.80 & 39.48 \\
MM-Eureka-Qwen-7B \citep{mmeureka} & 72.60 & 32.10 & 45.40 & 23.00 & 21.80 & 38.98 \\
InternVL3-8B \citep{internvl3} & 70.50 & 28.60 & 33.90 & 23.00 & 37.50 & 38.70 \\
OpenVLThinker-7B \citep{openvlthinker} & 65.30 & 26.90 & 38.10 & 16.80 & 35.20 & 36.46 \\
Qwen2.5-VL-7B-Instruct \citep{qwen25vl} & 69.20 & 21.80 & 34.10 & 18.00 & 32.30 & 35.08 \\
\midrule
\multicolumn{7}{c}{\emph{Supervised Fine-Tuning (ours).}} \\[1pt]
\rowcolor{gray!7} Seed-only SFT & 74.10 & 36.16 & 65.02 & 26.96 & 56.76 & 51.80 \\
\rowcolor{gray!7} \textbf{\method-SFT} (VeriEvol-SFT-Init) & 76.60 & 39.80 & 67.01 & 30.94 & 59.33 & 54.73 \\
\midrule
\multicolumn{7}{c}{\emph{Reinforcement Learning (ours).}} \\[1pt]
\rowcolor{gray!7} RL-Origin & 77.00 & 41.45 & 69.04 & 30.54 & 58.19 & 55.24 \\
\rowcolor{gray!7} RL-Evol & 78.10 & 45.39 & 69.67 & 32.14 & 60.00 & 57.06 \\
\rowcolor{gray!15} \textbf{RL-Evol + Verifier} (full \method) & 79.00 & \textbf{47.37} & \textbf{70.46} & \textbf{35.73} & \textbf{63.05} & \textbf{59.12} \\
\bottomrule
\end{tabular}}
\caption{Main results on five visual math reasoning benchmarks. External rows are paper-reported or comparison-table numbers; ours rows (SFT and RL) train Qwen2.5-VL-7B-Instruct under our harness (\S\ref{sec:exp_setup}). \textbf{Bold} marks the per-column best. Blank cells mark benchmarks a paper does not report under our metric; rows covering $\le 3/5$ benchmarks are shaded lighter and excluded from the $\mathrm{Avg.}$-column bold rule. $^\dagger$ marks a MathVerse split not confirmed as Vision-Only (excluded from the MathVerse-VO column).}
\label{tab:mainresults}
\end{table}
Table~\ref{tab:mainresults} answers two internal questions and places the answers in external context. Internally, the SFT rows ask whether evolution improves SFT-stage data, and the RL rows ask whether evolution and the verifier each add a separable RL-stage gain on a shared init. The external baseline rows place the resulting numbers against 7B-size baselines.

\paragraph{SFT-stage evidence.}
\method-SFT (VeriEvol-SFT-Init) lifts the five-benchmark mean from 51.80 to 54.73 over a Seed-only SFT baseline ($+2.93$), with consistent per-benchmark gains and a peak of $+3.98$ on DynaMath. Because the \method-SFT row differs from the seed baseline only in whether the SFT corpus is evolved, this delta is a clean isolation of the SFT-stage contribution of evolution and confirms that the operators add difficulty that survives SFT before any RL is applied.

\paragraph{RL-stage evidence.}
On top of VeriEvol-SFT-Init, data construction yields a clean stepwise improvement on every benchmark. Evolution alone (RL-Origin~$\to$~RL-Evol) raises the average from 55.24 to 57.06, peaking on MathVision\_MINI ($+3.94$) and We-Math ($+1.81$). Adding the \agent\ verifier (RL-Evol~$\to$~RL-Evol + Verifier) adds a further $+2.06$ on the same average and lifts every benchmark, peaking on DynaMath ($+3.59$) and We-Math ($+3.05$). Cumulatively, full \method\ improves over RL-Origin by $+3.88$ on average and $+5.92$ on MathVision\_MINI. Because all three RL rows share VeriEvol-SFT-Init and the GRPO recipe, these deltas isolate the \emph{marginal} contribution of evolved prompts and the verifier check at the RL stage; the SFT-stage component is bounded by the SFT rows (Section~\ref{sec:exp_setup}).

\begin{figure}[t]
\centering
\includegraphics[width=0.62\linewidth]{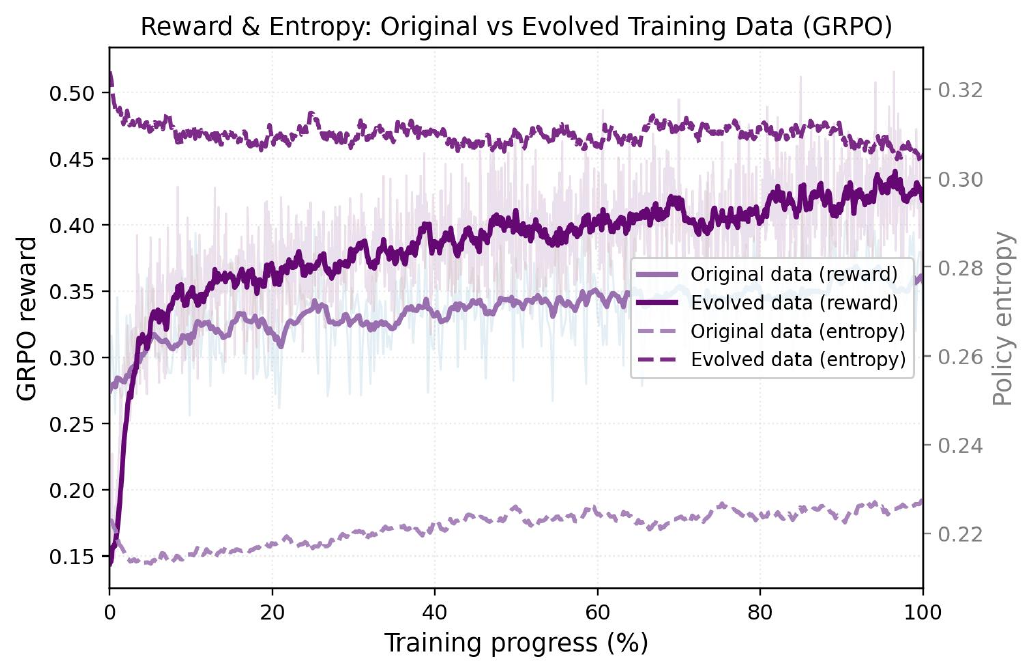}
\caption{\textbf{GRPO training dynamics on original vs.\ evolved prompts.} Solid curves are the GRPO reward (left axis); dashed curves are the policy entropy (right axis); the $x$-axis is normalized training progress. Light traces are the raw per-step signal, bold traces the running mean. Both runs share VeriEvol-SFT-Init, the GRPO recipe, and the verifier, differing only in whether the RL prompt pool is evolved (RL-Evol) or not (RL-Origin).}
\label{fig:grpo_dynamics}
\end{figure}

\paragraph{Training dynamics: why evolved prompts help.}
Figure~\ref{fig:grpo_dynamics} traces the GRPO reward and policy entropy for the RL-Origin and RL-Evol runs, which differ only in whether the prompt pool is evolved, and explains the $+1.82$ RL-stage gain mechanistically rather than only as an endpoint accuracy. Two effects are visible. First, on the reward axis the evolved-prompt run climbs faster early and settles at a markedly higher terminal reward (running mean $\approx 0.43$ vs.\ $\approx 0.35$ for original prompts), indicating that the evolution operators raise difficulty into a band that still yields a dense, learnable advantage signal rather than saturating the policy. Second, and more diagnostically, the evolved-prompt run \emph{sustains} a higher policy entropy throughout training ($\approx 0.31$, roughly flat) whereas the original-prompt run collapses to and then only slowly recovers from a much lower entropy ($\approx 0.22$). Low-difficulty seeds are solved by a narrow set of trajectories, so GRPO drives the policy toward a low-entropy mode early; the harder, image-grounded evolved prompts keep multiple reasoning routes viable and thus preserve exploration as reward rises. The combination---higher reward without entropy collapse---is the training-time signature of the evolved data and is consistent with RL-Evol's broad, non-degrading per-benchmark gains.

\paragraph{External comparison: leading on MathVerse-VO.}
On the four conventional splits \method\ is competitive but not uniformly best (detailed below); we therefore anchor the headline external comparison on MathVerse-VO, the only split that directly isolates image reliance and thus the most diagnostic test of what \method\ targets. Among external baselines, OpenMMReasoner-7B reports the strongest MathVerse-VO score (63.80), followed by MMR1-Math-v0 (55.40) and OVR-7B (54.60); against these competitors, our 70.46 leads by $+6.66$, $+15.06$, and $+15.86$ points respectively.

\paragraph{External comparison: competitive but not uniformly leading on the other four benchmarks.}
OpenMMReasoner-7B reaches 79.50 on MathVista\_MINI (vs.\ our 79.00) using a roughly $3.5\times$ larger SFT corpus (874K vs.\ our 250K), and ReVisual-R1-7B reaches 48.80 on MathVision\_MINI (vs.\ our 47.37). On DynaMath-Worst, full \method\ (35.73) leads every full-coverage external baseline (next-best OpenMMReasoner-7B, 34.90). Under the unified We-Math-Strict column, full \method\ reaches 63.05 and leads the strongest external strict score (OpenMMReasoner-7B, 53.81, which we reproduce under We-Math-Strict since its paper reports only the loose metric).

Cross-row averages in the $\mathrm{Avg.}$ column should be interpreted with care: some external systems report only a subset of the five benchmarks while our rows cover all five, so a higher external average can reflect averaging over a different subset.

\subsection{Scaling Analysis and Ablations}
\label{sec:exp_ablation}

We separate the scaling behavior of \method\ into data volume at the SFT stage, verified data volume at the RL stage, verifier modularity, and evolution depth at a fixed budget.

\begin{table}[!htbp]
\centering
\small
\setlength{\tabcolsep}{5pt}
\renewcommand{\arraystretch}{1.10}
\resizebox{\linewidth}{!}{%
\begin{tabular}{lcccccc}
\toprule
\# SFT data & \shortstack{MathVista\\Mini\,$\uparrow$} & \shortstack{MathVision\\Mini\,$\uparrow$} & \shortstack{MathVerse\\VO\,$\uparrow$} & DynaMath\,$\uparrow$ & We-Math\,$\uparrow$ & Avg.\,$\uparrow$ \\
\midrule
\phantom{0}10K & 66.00 & 20.39 & 34.58 & 19.05 & 37.06 & 35.42 \\
\phantom{0}50K & 68.80 & 23.38 & 38.80 & 21.05 & 42.45 & 38.90 \\
100K & 72.20 & 28.64 & 44.89 & 24.90 & 46.40 & 43.41 \\
150K & 74.30 & 32.68 & 54.28 & 28.00 & 51.75 & 48.20 \\
200K & 75.30 & 37.93 & 62.25 & 29.43 & 56.29 & 52.24 \\
\midrule
250K (\textbf{VeriEvol-SFT}) & \textbf{76.60} & \textbf{39.80} & \textbf{67.01} & \textbf{30.94} & \textbf{59.33} & \textbf{54.73} \\
\bottomrule
\end{tabular}}
\caption{Scaling evolved SFT data volume. All rows fine-tune Qwen2.5-VL-7B-Instruct with the same SFT recipe and differ only in the number of \sftdata\ samples. The 250K row is VeriEvol-SFT-Init from Table~\ref{tab:mainresults}.}
\label{tab:sft_scaling}
\end{table}

\paragraph{Scaling Axis 1: SFT data volume.}
Table~\ref{tab:sft_scaling} isolates data volume before any RL update. The five-benchmark mean increases monotonically from 35.42 at 10K to 54.73 at 250K ($+19.31$), showing that the evolved SFT corpus remains highly useful well beyond the small-data regime and justifying the 250K VeriEvol-SFT-Init used by all RL runs. The gains are largest on benchmarks that require stronger visual-compositional coverage: MathVerse-VO rises by $+32.43$, We-Math by $+22.27$, and MathVision\_MINI by $+19.41$. MathVista\_MINI and DynaMath improve more moderately ($+10.60$ and $+11.89$), with DynaMath rising only modestly between 10K and 50K ($+2.00$), suggesting that these benchmarks either saturate earlier under SFT or depend less on additional supervised coverage. Thus SFT scaling mainly builds the broad visual-reasoning capability of the initialization, while the verified RL sweep below tests whether reward-aligned training still benefits from additional verified samples on top of that initialization.

\begin{table}[!htbp]
\centering
\small
\setlength{\tabcolsep}{5pt}
\renewcommand{\arraystretch}{1.10}
\resizebox{\linewidth}{!}{%
\begin{tabular}{lcccccc}
\toprule
\# RL data & \shortstack{MathVista\\Mini\,$\uparrow$} & \shortstack{MathVision\\Mini\,$\uparrow$} & \shortstack{MathVerse\\VO\,$\uparrow$} & DynaMath\,$\uparrow$ & We-Math\,$\uparrow$ & Avg.\,$\uparrow$ \\
\midrule
\phantom{0}10K & 76.60 & 38.49 & 67.26 & 31.04 & 59.19 & 54.52 \\
\phantom{0}33K & 78.10 & 38.16 & 68.02 & 31.98 & 60.76 & 55.40 \\
\phantom{0}67K & 77.50 & 41.45 & 68.91 & 33.54 & 61.14 & 56.51 \\
100K & 78.10 & 45.08 & 69.29 & 34.24 & 61.29 & 57.60 \\
\midrule
130K (full \method) & \textbf{79.00} & \textbf{47.37} & \textbf{70.46} & \textbf{35.73} & \textbf{63.05} & \textbf{59.12} \\
\bottomrule
\end{tabular}}
\caption{Scaling verified RL data volume. All rows share VeriEvol-SFT-Init, the same GRPO recipe, and the end-of-one-epoch checkpoint on the corresponding subset of a single verified evolved-prompt pool; they differ only in sample count. The 130K row is the full RL-Evol + Verifier system from Table~\ref{tab:mainresults}.}
\label{tab:ablation}
\end{table}

\paragraph{Scaling Axis 2: verified RL data volume.}
Table~\ref{tab:ablation} studies the second scaling axis by varying only the number of verified RL samples while holding the SFT init, GRPO recipe, verifier model, and one-epoch training budget fixed. The five-benchmark mean improves smoothly from 54.52 at 10K to 59.12 at 130K ($+4.60$ across an order-of-magnitude scaling), showing that the pipeline does not merely produce a high-quality final pool but continues to supply usable reward signal as the pool grows. The improvement is far from uniform: MathVision\_MINI is the most data-hungry ($38.49 \to 47.37$, $+8.88$), MathVerse-VO and We-Math show smaller positive scaling ($+3.20$ and $+3.86$), and MathVista\_MINI scales least ($76.60 \to 79.00$, $+2.40$), which we read as near saturation at the backbone capacity. No benchmark worsens with scale, indicating that the verifier filter remains usable across the entire range rather than introducing a systematic bias on any single benchmark.

\begin{figure}[t]
\centering
\includegraphics[width=\linewidth]{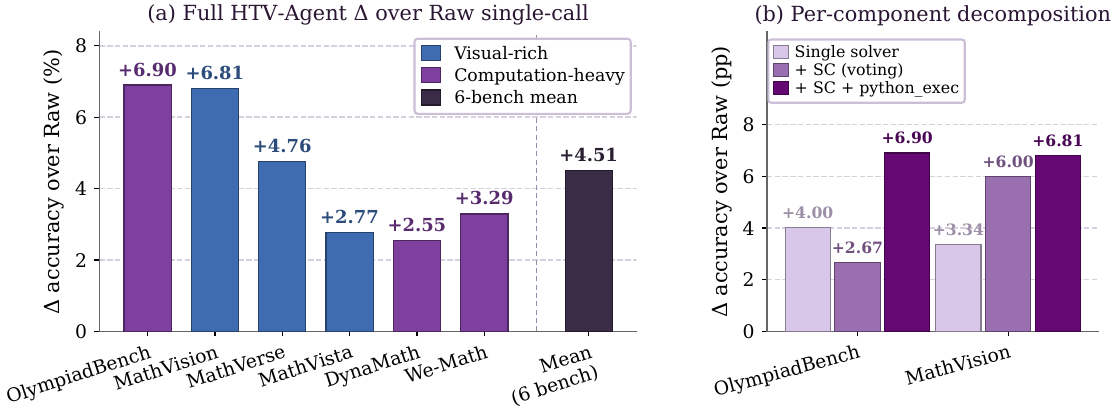}
\caption{\textbf{\agent\ component ablation as an inference-time judge}
(backbone Gemini-3.5-Flash, 300 samples per benchmark).
\textbf{(a)} $\Delta$ accuracy (pp) of the full \agent\
(SC\,+\,\texttt{python\_exec}) over the \emph{Raw} single-call baseline
on the five main benchmarks plus OlympiadBench and their mean; bars are
color-coded by benchmark family (visual-rich vs.\ computation-heavy).
\textbf{(b)} Per-component decomposition (Raw, Single solver, SC alone,
SC\,+\,\texttt{python\_exec}) on the two benchmarks where all four
configurations were run. Full decomposition and latency in
Appendix~\ref{app:agent_ablation}.}
\label{fig:agent_ablation}
\end{figure}

\paragraph{Scaling Axis 3: verifier-channel modularity.}
End-to-end RL training is expensive, so we cannot directly run a verifier on/off ablation at matched RL data budget (Limitation~ii.a). We instead probe whether \agent\ behaves as an extensible reliability layer by deploying it as an \emph{inference-time judge} on six multimodal math benchmarks (the five main benchmarks plus OlympiadBench \citep{olympiadbench}, using Gemini-3.5-Flash as the backbone, 300 samples per benchmark), and ablating its three core components---multi-solver self-consistency, the python\_exec tool channel, and the full HTV verifier---by cumulatively enabling them across four configurations: \textbf{Raw} (a single LLM call, no agent), \textbf{Single Solver} (the agent pipeline with one solver, no self-consistency voting), \textbf{SC} (three solvers with self-consistency voting, no \texttt{python\_exec}), and \textbf{SC\,+\,\texttt{python\_exec}} (the full \agent). Raw is the reference baseline for all reported $\Delta$ values; full per-configuration definitions are in Appendix~\ref{app:agent_ablation}. This capability-focused ablation uses the stronger Gemini-3.5-Flash backbone, distinct from the more economical Gemini-3-flash teacher used for large-scale \rldata\ construction in \S\ref{sec:exp_setup}; the difference is a deliberate cost trade-off (the cheaper model is used at construction scale, the stronger model for this small 300-sample-per-benchmark probe) and the ablation isolates the component design rather than the specific backbone. The reasoning is that if different evidence channels reliably improve raw single-call accuracy in complementary regimes, then the same modular channels are credible as offline filters when constructing verified RL data. Figure~\ref{fig:agent_ablation} reports these configurations.

Three patterns emerge from Figure~\ref{fig:agent_ablation}. The \emph{Raw} single-call baseline averages 80.89 over the listed suite (OlympiadBench 63.00, MathVision 83.33, MathVerse 70.33, MathVista 87.67, DynaMath 87.00, We-Math 94.00). First, the full \agent\ (SC + \texttt{python\_exec}) improves this mean by $+4.51$pp and never degrades any benchmark. Second, multi-solver self-consistency is the dominant component on visually rich tasks: MathVision improves by $+6.00$pp from voting alone, while \texttt{python\_exec} adds only $+0.81$pp on top because the problems are visual rather than computational. Third, \texttt{python\_exec} is the dominant component on computation-heavy OlympiadBench ($+4.23$pp on top of SC), confirming that the programmatic and visual evidence channels in \agent\ are complementary rather than redundant. This complementarity is the key extensibility property: new verifier channels should be added when they target failure modes not covered by existing channels, rather than as undifferentiated extra judging. Because the same evidence channels run inside \agent\ when it screens candidate answers during data construction, these inference-time gains are a credible indirect indicator of \agent's RL-data verification quality, even though they do not substitute for the matched-budget RL ablation that Limitation~ii.a documents.

\paragraph{Scaling Axis 4: evolution depth at fixed budget.}
The data-volume sweeps above vary \emph{how much} data the pipeline uses; we now hold volume fixed and vary \emph{how many evolution rounds} produced it, to test whether deeper evolution improves data quality independently of scale. Table~\ref{tab:ablation_rounds} compares one vs.\ two rounds of \method\ evolution at a matched 10K budget, under both pure SFT and GRPO RL, on the same five-benchmark suite. Two rounds improve every SFT benchmark (Avg.\ $35.42\to36.79$, $+1.37$), with the largest gains on MathVista\_MINI ($+2.42$) and MathVision\_MINI ($+1.57$); under RL the mean also improves ($54.52\to55.02$, $+0.50$), with all five benchmarks rising and We-Math nearly flat within noise. We read this as evidence that the evolution loop sharpens prompts in a way that is not captured by data volume alone: SFT, which fits the verified labels directly, is more sensitive to prompt quality, while RL at 10K is closer to saturation under our backbone and recipe and benefits more from additional RL volume (Axis~2) than from deeper evolution at fixed volume. Evolution depth and data volume are therefore complementary rather than substitutable.

\begin{table}[!htbp]
\centering
\small
\setlength{\tabcolsep}{5pt}
\renewcommand{\arraystretch}{1.10}
\resizebox{\linewidth}{!}{%
\begin{tabular}{llcccccc}
\toprule
Regime & Rounds & \shortstack{MathVista\\Mini\,$\uparrow$} & \shortstack{MathVision\\Mini\,$\uparrow$} & \shortstack{MathVerse\\VO\,$\uparrow$} & DynaMath\,$\uparrow$ & We-Math\,$\uparrow$ & Avg.\,$\uparrow$ \\
\midrule
\multirow{2}{*}{SFT} & 1 & 66.00 & 20.39 & 34.58 & 19.05 & 37.06 & 35.42 \\
                    & 2 & \textbf{68.42} & \textbf{21.96} & \textbf{35.24} & \textbf{20.54} & \textbf{37.80} & \textbf{36.79} \\
\addlinespace
\multirow{2}{*}{RL}  & 1 & 76.60 & 38.49 & 67.26 & 31.04 & 59.19 & 54.52 \\
                    & 2 & \textbf{77.10} & \textbf{39.21} & \textbf{67.54} & \textbf{31.92} & \textbf{59.35} & \textbf{55.02} \\
\bottomrule
\end{tabular}}
\caption{Evolution-depth ablation at a matched 10K data budget. ``Rounds'' is the number of \method\ evolution passes used to produce the training pool; all other factors (SFT init for the RL rows, GRPO recipe, verifier model, one-epoch budget) are held fixed within each regime. \textbf{Bold} marks the per-regime best.}
\label{tab:ablation_rounds}
\end{table}

\subsection{Error Analysis}
\label{sec:exp_error}

The benchmark numbers in Section~\ref{sec:exp_main} show that \method\ improves accuracy, but they do not by themselves show \emph{why}. To address this, we report a small \emph{pilot} audit on the two benchmarks in our suite that most directly test reliance on the image (MathVerse-VO and MathVision\_MINI), and analyze the same questions under RL-Origin and under RL-Evol + Verifier (the full method). The pilot is designed around three questions: (i) does \method\ \emph{redistribute} the model's errors away from visual grounding failures? (ii) do successful traces actually \emph{use} the image more? (iii) where does the full system still fail? All numbers below come from a 60-sample single-annotator pilot and are illustrative of qualitative trends. The three layers below address (i)--(iii) in turn.

\paragraph{Audit protocol.}
The pilot draws 60 samples (30 from MathVerse-VO and 30 from MathVision\_MINI, no problem family $>$35\% of the pool). We collect both models' final answer and full chain-of-thought under the same evaluation protocol as Section~\ref{sec:exp_setup} (one trace per problem), and a single annotator labels errors into the five-class taxonomy below. Taxonomy and pilot data are released with the code.

\paragraph{Layer 1: error-type distribution shift.}
We label each error into five mutually exclusive types: \textbf{(VG)} visual grounding, where the model misreads, ignores, or hallucinates an image element; \textbf{(RC)} reasoning chain, where visual content is parsed correctly but the multi-step argument breaks; \textbf{(SA)} symbolic/arithmetic step error; \textbf{(AF)} answer formatting; \textbf{(TS)} text-only shortcut. Table~\ref{tab:error_dist} reports the joint distribution on the 60-sample pilot. Two trends are visible in the pilot, both reported as absolute counts to keep the small sample size legible. First, VG and TS errors drop from 16 to 9 in absolute terms (out of 28 vs.\ 24 total errors); their share of the residual error pool also drops from 57\% to 38\%, suggesting that residual errors shift away from the categories \method\ directly targets. Second, the absolute drop in VG errors ($-3$ on MathVerse-VO, $-4$ on MathVision\_MINI, $-7$ in total) is the largest single-category drop in the pilot, in the same direction as the $+1.42$ MathVerse-VO and $+5.92$ MathVision\_MINI gains in Table~\ref{tab:mainresults}. RC and SA counts are essentially flat or slightly higher in absolute terms ($2{\to}2$, $1{\to}2$ on MathVerse-VO; $4{\to}5$, $3{\to}4$ on MathVision\_MINI), so in a smaller error pool they become a \emph{larger share} of the residual; this is the backbone-capacity pattern we describe under Layer~3 (R2). At $n{=}60$ with a single annotator we draw no quantitative conclusion from these proportions.

\begin{table}[!htbp]
\centering
\footnotesize
\setlength{\tabcolsep}{5pt}
\renewcommand{\arraystretch}{1.05}
\begin{tabular}{lcccccc}
\toprule
& \multicolumn{5}{c}{Errors per category (out of 30, $\downarrow$)} & \\
\cmidrule(lr){2-6}
Setting (per benchmark slice) & VG & RC & SA & AF & TS & Total errors\,$\downarrow$ \\
\midrule
\multicolumn{7}{l}{\textit{MathVerse-VO (30 audited)}} \\
RL-Origin & 5 & 2 & 1 & 1 & 0 & 9 \\
RL-Evol + Verifier (full) & 2 & 2 & 2 & 1 & 1 & 8 \\
\addlinespace
\multicolumn{7}{l}{\textit{MathVision\_MINI (30 audited)}} \\
RL-Origin & 8 & 4 & 3 & 1 & 3 & 19 \\
RL-Evol + Verifier (full) & 4 & 5 & 4 & 1 & 2 & 16 \\
\bottomrule
\end{tabular}
\caption{\textbf{Pilot} error-type distribution from a 60-sample audit (30 per benchmark) by a single annotator. VG = visual grounding, RC = reasoning chain, SA = symbolic/arithmetic, AF = answer formatting, TS = text-only shortcut. Counts are illustrative of the qualitative pattern and are not significance-tested.}
\label{tab:error_dist}
\end{table}

\paragraph{Layer 2: image use in successful traces.}
Aggregate error counts are necessary but not sufficient: a system can score higher while still relying on text-only patterns. To check whether the Layer-1 shift is accompanied by qualitatively different traces, we extract the pilot problems on which RL-Origin fails and RL-Evol + Verifier succeeds ($n{=}5$ on MathVerse-VO; $n{=}7$ on MathVision\_MINI; $n{=}12$ in total) and score each successful trace, paired against the corresponding failed RL-Origin trace on the same problem, on three binary criteria: explicit image reference, image-grounded subgoal decomposition, and answer consistency with image content (vs.\ text-only priors). Across the 12 pairs, RL-Evol + Verifier satisfies the first criterion on 10/12 ($83\%$) of its successful traces versus 5/12 ($42\%$) for the matched failed RL-Origin traces, with similar gaps on the other two criteria (9/12 vs.\ 3/12; 11/12 vs.\ 5/12). Two case studies in Appendix~\ref{app:cases} illustrate the shift; statistical significance at this $n$ is not claimed.

\paragraph{Layer 3: residual failure modes.}
\method\ still fails along two patterns observed in the pilot. \textbf{(R1)} \emph{Solver--verifier shared blind spots}: roughly 3 of the 60 audited problems ($\sim$5\%) where all three solver branches and the verifier converge on the same incorrect visual reading and the gates accept it. These are the cases where the \agent\ "refute-not-confirm" framing (\S\ref{sec:agents}) reduces to "confirm" because every channel makes the same mistake; a heterogeneous verifier ensemble (different backbone, different tool stack) is the natural remedy. \textbf{(R2)} \emph{Backbone-capacity ceiling}: long-tail MathVision\_MINI items requiring multi-step symbolic chains beyond the 7B backbone's reliable reasoning length, manifesting as RC/SA errors largely insensitive to the data framework (these account for 12 of the 60 audited problems ($\sim$20\%) and explain the flat-or-rising RC/SA columns in Table~\ref{tab:error_dist}); a larger backbone is the natural remedy.

\section{Limitations, Broader Impact, and Responsible Release}
\label{sec:limits}

\paragraph{Limitations.}
\method\ depends on seed-pool coverage: evolution cannot fully compensate for under-represented domains or diagram styles, and the gates, though conservative, are imperfect---evolved prompts may still leak textual cues, and \agent\ can accept a wrong answer when solvers, verifiers, and tools share a blind spot. The RL-ready subset is therefore necessarily narrower than the SFT-ready subset. On the empirical side, each RL run uses a single seed, so single-benchmark differences below $\sim$1 point should be read as suggestive rather than significant, and the verifier-on/off and SFT-stage-evolution effects are isolated only indirectly through the SFT and RL rows of Table~\ref{tab:mainresults} rather than through fully matched controls. These are scoping choices that bound the granularity of our claims, not threats to the main scaling trends.

\paragraph{Positive impact, risks, and safeguards.}
The framework can reduce the cost of building high-quality training data for visual math reasoning and make the construction process transparent for smaller teams. Synthetic pipelines can amplify source bias, overfit to generator/judge preferences, or create spurious reward shortcuts; raw images with restrictive licenses or privacy concerns pose legal/ethical risks. We release per-source license tracking, dedup, and content filtering, and substitute metadata plus source pointers for raw images when redistribution rights are unclear.

\section{Conclusion}
\label{sec:conclusion}
\vspace{-0.5em}

We framed RL data construction for visual mathematical reasoning as a scaling problem: prompt difficulty must grow without losing image dependence, answer labels must remain reliable as the pool expands, and the resulting supervision must plug into existing policy optimizers. \method\ addresses these requirements by decoupling route-specific prompt evolution from hypothesis--test answer falsification before any policy update.

At the SFT stage, scaling evolved data from 10K to 250K raises the five-benchmark mean by $+19.31$; with backbone, SFT init, and GRPO recipe held fixed, the RL-stage design further improves the mean by $+3.88$ over an un-evolved RL baseline ($+5.92$ on MathVision\_MINI) and continues to scale from 10K to 130K verified RL samples ($+4.60$). The component analysis supports another scaling axis: verifier channels are complementary rather than redundant, so reliability can be extended by adding targeted refutation channels instead of redesigning the optimizer.

The broader takeaway is composability. Data-stage verification is orthogonal to policy-side innovations, allowing future GRPO-style optimizers, length penalties, or rollout reweighting strategies to build on \rldata\ directly; it also suggests extensions beyond mathematical reasoning to chart QA, diagram-based science, and document understanding, wherever image-grounded answers admit deterministic or programmatic checks. Code, data, models, and verifier traces will be released so that downstream work can audit, extend, and rescale the construction pipeline.


\FloatBarrier
\appendix

\newpage

\section{Training Compute and Hyperparameters}
\label{app:compute}

This appendix documents the training configuration of \method\ across the SFT and RL stages. Hyperparameter values reported here are consistent with the description in the main text (Section~\ref{sec:exp_setup}). We specify the hardware family and the deterministic configuration manifest under which every reported run can be reproduced.

\paragraph{SFT stage (VeriEvol-SFT-Init).}
We fine-tune Qwen2.5-VL-7B-Instruct on the 250K-sample \sftdata\ corpus to obtain VeriEvol-SFT-Init. We use sequence packing with a global batch size of 512, AdamW with a constant learning rate of $5\times10^{-5}$, and train for 10 epochs. The resulting checkpoint serves as the starting point for every RL run reported in the paper.

\paragraph{RL stage (GRPO on \rldata).}
All RL variants share the GRPO recipe described in Section~\ref{sec:exp_setup}: global batch size 256, group size 16, dynamic batching, and a binary reward of $1$ when the model's extracted answer matches the verified reference under the verifier $v$ and $0$ otherwise (with a small format-error penalty $\beta$). The same prompt format and the same reference-answer extraction routine are used across RL-Origin, RL-Evol, RL-Evol + Verifier, and the four data-scale ablation runs.

\paragraph{Compute and reproducibility.}
All SFT and RL runs were carried out on a homogeneous H20 (80\,GB) GPU cluster, with each run using a single 8-GPU node and tensor-parallelism degree 8; the SFT run additionally uses sequence-packed data parallelism across two such nodes. As an order-of-magnitude indication of the total compute, the reported experiments comprise one 250K-sample SFT run plus 8 GRPO runs (RL-Origin, RL-Evol, RL-Evol+Verifier, the four \rldata\ scale points at 10/33/67/100/130K with 130K shared, and the round-2 ablation), each on the same 8$\times$H20 node. For exact reproduction we release: (i) a single configuration manifest (YAML) for the SFT and the 130K RL run that pins the framework version, optimizer, scheduler, batch dimensions, group size, KL coefficient, response-length budget, and seed; and (ii) the full \rldata\ and \sftdata\ artifacts and verifier traces, so that any third party can rerun the released configurations on equivalent hardware.

\section{HTV-Agent Component Ablation: Full Details}
\label{app:agent_ablation}

This appendix expands Figure~\ref{fig:agent_ablation} in the main text. We deploy \agent\ as an inference-time judge on the five main multimodal math benchmarks plus OlympiadBench and ablate its three core components (multi-solver self-consistency, programmatic execution via the \texttt{python\_exec} tool channel, and the verifier filter that decides whether to accept or fall back to the raw answer). The goal is not to claim a new state-of-the-art at inference time, but to provide indirect evidence that the same component design is a reliable filter when used offline to verify candidate RL training data.

\paragraph{Setup.}
The backbone for every configuration is Gemini-3.5-Flash, accessed through a shared API gateway. Each benchmark draws 300 samples. Common high-effort settings: \texttt{reasoning\_effort=high}, \texttt{max\_tokens=8192}, \texttt{max\_solver\_steps=16}, \texttt{max\_verifier\_steps=10}, per-sample timeout 600\,s. Four configurations are compared.

\begin{itemize}
\item \textbf{Raw Baseline.} A single LLM call, no agent. Direct prompt-to-answer.
\item \textbf{Single Solver.} The agent pipeline with one solver (no self-consistency voting), with \texttt{python\_exec} enabled on computation-related benchmarks and the verifier filter active.
\item \textbf{SC (no python\_exec).} Three solvers with self-consistency voting, but the \texttt{python\_exec} tool channel is disabled; the verifier filter is active.
\item \textbf{SC + python\_exec.} The full \agent\ pipeline with three solvers, self-consistency voting, \texttt{python\_exec} tool, and the verifier filter.
\end{itemize}

\paragraph{Full agent-ablation accuracy table.}
Table~\ref{tab:agent_ablation_full} reports the absolute accuracy of each
configuration on the five main benchmarks plus OlympiadBench.
Figure~\ref{fig:agent_ablation} in the main text plots the
$\Delta$-over-Raw view of the same data; this table provides the absolute
numbers for reproducibility, including the two configurations (Single
Solver and Raw) that were run on the full listed suite and the SC-only
configuration that was run only on OlympiadBench and MathVision (see
Notes on data validity below).

\begin{table}[h]
\centering
\small
\setlength{\tabcolsep}{4pt}
\renewcommand{\arraystretch}{1.10}
\resizebox{\linewidth}{!}{%
\begin{tabular}{lccccc}
\toprule
Benchmark & Raw\,$\uparrow$ & Single Solver\,$\uparrow$ & SC (no exec)\,$\uparrow$ & SC + \texttt{python\_exec}\,$\uparrow$ & Best $\Delta$ \\
\midrule
OlympiadBench & 63.00 & 67.00 & 65.67 & \textbf{69.90} & \textbf{+6.90} \\
MathVision & 83.33 & 86.67 & 89.33 & \textbf{90.14} & \textbf{+6.81} \\
MathVerse & 70.33 & 72.33 & -- & \textbf{75.09} & \textbf{+4.76} \\
MathVista & 87.67 & 88.33 & -- & \textbf{90.44} & \textbf{+2.77} \\
DynaMath & 87.00 & 87.67 & -- & \textbf{89.55} & +2.55 \\
We-Math & 94.00 & 96.67 & -- & \textbf{97.29} & +3.29 \\
\midrule
Mean (listed suite) & 80.89 & 83.11 & -- & \textbf{85.40} & \textbf{+4.51} \\
\bottomrule
\end{tabular}}
\caption{\textbf{Full \agent\ component ablation as an inference-time
judge: absolute accuracies on the five main benchmarks plus
OlympiadBench} (backbone Gemini-3.5-Flash, 300 samples per benchmark;
absolute-value counterpart of Figure~\ref{fig:agent_ablation}). ``--''
marks configurations not run on that benchmark (the SC-only
configuration was a targeted probe on OlympiadBench and MathVision; see
Notes on data validity). \textbf{Bold} marks the per-benchmark best;
\emph{Best $\Delta$} reports the gap to Raw. Within-row Mean covers only
the benchmarks each row runs.}
\label{tab:agent_ablation_full}
\end{table}

\paragraph{Per-component decomposition: isolating the role of self-consistency.}
Table~\ref{tab:agent_decomp} re-examines the two benchmarks where all
four configurations were run, with the explicit goal of isolating
\emph{how much of the full-pipeline gain is already explained by
self-consistency voting alone}. We highlight the SC-only row because it
shows that on visually rich MathVision, three-solver voting recovers
$+6.00$pp without any code execution---i.e., SC alone is the dominant
component on that benchmark, and \texttt{python\_exec} adds only
$+0.81$pp further. On computation-heavy OlympiadBench the picture inverts: SC
alone only adds $+2.67$pp, while adding \texttt{python\_exec} on top of
SC delivers an additional $+4.23$pp, making the programmatic channel
the dominant component there. This per-benchmark inversion is the
clearest evidence we have that the two evidence channels are
complementary rather than redundant.

\begin{table}[h]
\centering
\small
\setlength{\tabcolsep}{6pt}
\renewcommand{\arraystretch}{1.10}
\begin{tabular}{lccc}
\toprule
& \multicolumn{3}{c}{Accuracy ($\uparrow$) ; ($\Delta$ over Raw)} \\
\cmidrule(lr){2-4}
Configuration & OlympiadBench & MathVision & Mean of 2 \\
\midrule
Raw Baseline & 63.00 & 83.33 & 73.17 \\
\midrule
+ Single Solver (no voting) & 67.00 ($+4.00$) & 86.67 ($+3.34$) & 76.84 ($+3.67$) \\
\rowcolor{gray!12}
+ Self-consistency voting \emph{(no exec)} & 65.67 ($+2.67$) & 89.33 ($+6.00$) & 77.50 ($+4.33$) \\
+ \texttt{python\_exec} on top of SC & \textbf{69.90 ($+6.90$)} & \textbf{90.14 ($+6.81$)} & \textbf{80.02 ($+6.86$)} \\
\bottomrule
\end{tabular}
\caption{\textbf{Per-component decomposition on the two benchmarks
where all four configurations were run} (300 samples each).
\emph{Shaded row}: self-consistency voting alone (without
\texttt{python\_exec}). Values in parentheses are $\Delta$ over Raw;
\textbf{bold} marks the per-column best.}
\label{tab:agent_decomp}
\end{table}

\paragraph{Latency.}
The full \agent\ pipeline is substantially more expensive than a single LLM call because it runs three solver passes plus the verifier. Table~\ref{tab:agent_latency} reports per-sample wall-clock cost. This is the price for inference-time gains, but at the data-construction stage the cost is amortized over the lifetime of the verified \rldata\ pool; it is paid once and reused across every downstream RL run.

\begin{table}[h]
\centering
\small
\setlength{\tabcolsep}{8pt}
\renewcommand{\arraystretch}{1.10}
\begin{tabular}{lcc}
\toprule
Configuration & Latency per sample & Relative to Raw \\
\midrule
Raw Baseline & 14--35\,s & 1$\times$ \\
Single Solver & 55--97\,s & 4--6$\times$ \\
SC (no \texttt{python\_exec}) & 367--433\,s & 12--25$\times$ \\
SC + \texttt{python\_exec} & 344--465\,s & 15--20$\times$ \\
\bottomrule
\end{tabular}
\caption{Inference-time latency per sample, relative to a single Raw API call. At the data-construction stage this cost is paid once per training sample and reused across every RL run that consumes \rldata.}
\label{tab:agent_latency}
\end{table}

\paragraph{Key qualitative findings.}
Three observations summarize the table. (1) The full \agent\ never degrades the listed-suite mean and never drops any single benchmark below the raw baseline by more than the validity caveats listed below; the multi-solver voting plus a verifier filter that can abstain back to the raw answer act as a safety net. (2) The two evidence channels (multi-solver voting versus programmatic execution) are complementary: voting helps most when reasoning paths benefit from diversity, while \texttt{python\_exec} helps most when problems involve explicit arithmetic. (3) The agent pipeline had zero pipeline-level errors across all five benchmarks in our runs, while the raw baseline occasionally suffered timeout or connection errors; this robustness is a side benefit of the multi-solver retry architecture and is one of the reasons the same architecture is a reliable offline filter for \rldata\ construction.

\paragraph{Notes on data validity.}
Because we ran multiple configurations across two seeds and two API channels during the ablation campaign, several caveats apply. \textbf{(i)} The SC and \texttt{python\_exec}-related configurations were run on seed 7, while the v3 single-solver and SC-only ablations were run on seed 42. Cross-seed comparisons are therefore approximate. \textbf{(ii)} A small number of SC runs were terminated before completing all 300 samples (e.g., SC on OlympiadBench completed 299/300 because the last sample was killed; SC on MathVision had 213/300 valid pairs after API budget exhaustion; the SC + \texttt{python\_exec} run on We-Math reports the score over $255/300$ valid samples after API budget exhaustion); the table reports each benchmark's accuracy over the largest available valid subset. \textbf{(iii)} Two API channels (\texttt{api\_google} and \texttt{api\_naci}) were used; both route to the same underlying Gemini-3.5-Flash model.

\section{Evolution Operator Design Space and the 12-Topic Instantiation}
\label{app:operators}

This appendix presents the design space we considered when prototyping \method, organized along four axes: the visual route, the reasoning skill to amplify, the answer type, and the verification contract. We list the full taxonomy here for two reasons. First, the operator design is itself a contribution: it documents the kinds of constraints a multimodal evol-instruct framework has to respect to produce both SFT-usable and RL-usable data, and it serves as a practical checklist for adapting \method\ to new domains. Second, our current implementation is a deliberate restriction of this design space: rather than instantiating every cell of the four-axis cross product, the production pipeline collapses the routing axes into a 12-topic vocabulary (OCR, image description, detection, analysis, content creation, suggestions, summarization, logical reasoning, science, concept extraction, medical imaging, and scene understanding), and associates one evolution prompt template per topic. Each template bundles the relevant visual-context preservation, reasoning-difficulty operator, answer-type constraint, and verifier contract for that topic, rather than composing them at runtime. The taxonomy below should therefore be read as the upper bound of the design we explored, with the 12-topic instantiation being the subset that has been deployed and audited in this paper. We separate operators into the four axes mentioned above; a generation prompt selects one route, applies the route's reasoning operator, and attaches the route's answer-type operator with its corresponding verifier.

\begingroup
\footnotesize
\setlength{\tabcolsep}{3pt}
\renewcommand{\arraystretch}{1.25}
\setlength{\heavyrulewidth}{1.1pt}
\setlength{\lightrulewidth}{0.5pt}
\begin{longtable}{@{}>{\raggedright\arraybackslash}p{0.12\linewidth}
                    >{\raggedright\arraybackslash}p{0.15\linewidth}
                    >{\raggedright\arraybackslash}p{0.35\linewidth}
                    >{\raggedright\arraybackslash}p{0.30\linewidth}@{}}
\caption{Full visual-route-conditioned evolution operator taxonomy.}
\label{tab:app-visual-operators}\\
\toprule
\taxhead{Route} & \taxhead{Trigger evidence} & \taxhead{Evolution operators} & \taxhead{Verification / rejection contract} \\
\midrule
\endfirsthead
\midrule
\taxhead{Route} & \taxhead{Trigger evidence} & \taxhead{Evolution operators} & \taxhead{Verification / rejection contract} \\
\midrule
\endhead
\endfoot
\bottomrule
\endlastfoot
\taxrow{Geometry diagrams} & Points, lines, angles, polygons, circles, labels, congruence marks & Auxiliary construction\opsep relation composition\opsep theorem/invariant query\opsep coordinate conversion\opsep inverse solve from target value\opsep ratio, perimeter, area, or length comparison\opsep multi-step equation setup\opsep degenerate-case exclusion & Every symbol must bind to a visible mark\opsep reject impossible measurements, purely text-only theorem recall, and questions requiring precision not supported by the diagram \\
\taxsep
\taxrow{Coordinate geometry and function plots} & Axes, ticks, grids, curves, intercepts, shaded regions, plotted points & Graph-to-equation mapping\opsep intercept/extremum/intersection query\opsep slope or rate computation\opsep area/accumulation approximation\opsep parameter-shift reasoning\opsep symbolic-numeric consistency check\opsep interval selection & Require coordinate evidence\opsep normalize axis scales\opsep reject unseen precision, extrapolation outside visible support, and ambiguous intersections \\
\taxsep
\taxrow{Statistical charts} & Axes, legends, bars, lines, pies, scales, error bars & Cross-series comparison\opsep extrema/ranking\opsep aggregate over visible groups\opsep rate or slope change\opsep interpolation inside visible range\opsep unit/scale/log conversion\opsep conditional trend query\opsep uncertainty-aware comparison & Require cited axis/legend evidence\opsep reject chart-title-only questions, unsupported extrapolation, and ambiguous visual encodings \\
\taxsep
\taxrow{Tables and spreadsheets} & Rows, columns, headers, cells, subtotals, footnotes & Row-column join\opsep multi-cell aggregation\opsep conditional filtering\opsep subtotal or percentage computation\opsep rank/order query\opsep unit normalization across columns\opsep consistency check against totals & Require header-cell alignment\opsep reject hallucinated rows/columns, implicit database facts, and aggregation over hidden cells \\
\taxsep
\taxrow{OCR and document math} & Local text spans, formulas, captions, page layout, footnotes, boxed expressions & Multi-region lookup\opsep formula-text alignment\opsep equation extraction and solving\opsep distractor text suppression\opsep layout-conditioned aggregation\opsep cross-reference between caption, figure, and formula\opsep variable definition lookup & Require source spans or regions\opsep reject unsupported entities, ambiguous references, and questions answerable from OCR text without the relevant visual structure when image dependence is required \\
\taxsep
\taxrow{Maps, floorplans, and spatial layouts} & Coordinates, routes, rooms, grids, compass marks, scale bars, landmarks & Shortest/longest path\opsep direction and relative-position query\opsep scale conversion\opsep region inclusion/exclusion\opsep distance aggregation\opsep coordinate transform\opsep route constraint satisfaction & Require visible scale, grid, or landmark evidence\opsep reject unverifiable navigation/world-knowledge facts and missing-start/end-point prompts \\
\taxsep
\taxrow{Science process diagrams} & Arrows, states, labels, apparatus, biological\slash chemical\slash physical processes & Symbol-component binding\opsep process direction inference\opsep conservation or constraint application\opsep state-change comparison\opsep controlled-variable reasoning\opsep cause-effect chain extraction\opsep diagram-plus-equation coupling & Require figure-specific evidence\opsep reject pure domain trivia and mechanisms not represented in the diagram \\
\taxsep
\taxrow{Circuits, mechanics, and systems diagrams} & Components, wires, forces, pulleys, blocks, gears, flows, switches & Equivalent-structure reduction\opsep series/parallel or force-balance reasoning\opsep state toggling\opsep conservation constraints\opsep component removal/addition counterfactual\opsep input-output relation derivation & Require component-level grounding\opsep reject prompts whose answer depends on unstated physical parameters \\
\taxsep
\taxrow{Natural quantitative scenes} & Countable objects, spatial relations, occlusion cues, sizes, distances & Conditional counting\opsep grouping and selection\opsep size/distance comparison\opsep occlusion-aware count\opsep relation-chain reasoning\opsep scene-to-arithmetic conversion\opsep visible-ratio estimate & Require visible objects and relations\opsep reject subjective attributes, pure commonsense, and objects too small or occluded to verify \\
\taxsep
\taxrow{Visual patterns, grids, and matrices} & Repeated shapes, cells, transformations, colors, symmetry, missing entries & Pattern completion\opsep rotation/reflection/translation query\opsep symmetry or invariant detection\opsep combinatorial count\opsep row/column rule induction\opsep missing-cell reasoning\opsep transformation sequence inference & Require observable rule support\opsep reject puzzles with multiple valid completions unless additional constraints make the answer unique \\
\taxrow{Multi-panel and temporal images} & Ordered panels, before/after states, multiple views, sequence labels & Cross-panel delta\opsep temporal ordering\opsep causal state transition\opsep common/different attribute comparison\opsep aggregation across views\opsep counterfactual within observed state changes\opsep consistency across viewpoints & Require panel identity and ordering\opsep reject changes not visible between panels or questions needing external video context \\
\taxsep
\taxrow{Graphs, networks, and trees} & Nodes, edges, weights, arrows, hierarchies, adjacency, paths & Path length\opsep reachability\opsep degree/count query\opsep subgraph comparison\opsep hierarchy traversal\opsep flow or dependency reasoning\opsep tree-depth or least-common-ancestor query & Require visible node/edge labels\opsep reject missing edge weights, hidden adjacency, and graph-theoretic assumptions not stated visually \\
\taxsep
\taxrow{Venn, set, and probability diagrams} & Regions, overlaps, labels, counts, partitions, tree probabilities & Inclusion-exclusion\opsep conditional probability\opsep set difference/intersection\opsep missing-region solve\opsep consistency with total\opsep probability-tree path aggregation & Require visible counts/probabilities\opsep reject prompts with underdetermined region values or unstated independence assumptions \\
\taxsep
\taxrow{3D objects and perspective geometry} & Solids, faces, nets, projections, depth cues, rotations & Surface/volume computation\opsep net-to-solid mapping\opsep view transformation\opsep hidden-face inference when specified\opsep spatial relation comparison\opsep cross-section reasoning & Require visible dimensions or stated scale\opsep reject hidden measurements unless inferable from given constraints \\
\taxsep
\taxrow{Timelines and event sequences} & Time marks, durations, ordered events, intervals, dependencies & Duration aggregation\opsep overlap comparison\opsep before/after constraint solving\opsep rate over time\opsep missing event-time inference\opsep schedule conflict query & Require visible timestamps or intervals\opsep reject external historical facts and underdetermined schedules \\
\taxsep
\taxrow{Plots with uncertainty or measurement noise} & Error bars, confidence bands, approximate ticks, noisy observations & Bound-aware comparison\opsep interval overlap\opsep significant-difference query\opsep robust rank under uncertainty\opsep approximate numeric extraction with tolerance & Require tolerance or interval answer\opsep reject exact-value questions when visual uncertainty dominates \\
\end{longtable}
\endgroup

\subsection{Reasoning-Skill Operators}

\begingroup
\footnotesize
\setlength{\tabcolsep}{3pt}
\renewcommand{\arraystretch}{1.25}
\setlength{\heavyrulewidth}{1.1pt}
\setlength{\lightrulewidth}{0.5pt}
\begin{longtable}{@{}>{\raggedright\arraybackslash}p{0.17\linewidth}
                    >{\raggedright\arraybackslash}p{0.43\linewidth}
                    >{\raggedright\arraybackslash}p{0.33\linewidth}@{}}
\caption{Route-independent reasoning-skill operators.}
\label{tab:app-reasoning-operators}\\
\toprule
\taxhead{Operator} & \taxhead{Prompt transformation} & \taxhead{Acceptance signal} \\
\midrule
\endfirsthead
\midrule
\taxhead{Operator} & \taxhead{Prompt transformation} & \taxhead{Acceptance signal} \\
\midrule
\endhead
\endfoot
\bottomrule
\endlastfoot
\taxrow{Multi-hop composition} & Combine two or more visible facts before asking for the final answer & Intermediate facts can be cited and checked independently \\
\taxsep
\taxrow{Cross-region aggregation} & Require evidence from separated regions, panels, rows, cells, or diagram components & Evidence metadata contains multiple grounded regions \\
\taxsep
\taxrow{Constraint injection} & Add a condition that filters visible candidates or changes the computation path & Added condition is visually or textually supported and not contradictory \\
\taxsep
\taxrow{Inverse problem construction} & Ask for an input, parameter, or missing quantity that would produce a visible output & Solver can verify by substituting the answer forward \\
\taxsep
\taxrow{Comparative reasoning} & Convert direct reading into greater-than, difference, ratio, rank, or top-$k$ comparison & Candidate answer is uniquely ranked after normalization \\
\taxsep
\taxrow{Extrema search} & Ask for maximum, minimum, earliest, latest, nearest, farthest, or most/least frequent item & Search domain is finite and visible \\
\taxsep
\taxrow{Unit and scale conversion} & Require conversion across axis scales, map scales, table units, percentages, or dimensional units & Unit mapping is stated or visible\opsep normalized value passes tolerance \\
\taxsep
\taxrow{Rate and change reasoning} & Ask for slope, percent change, temporal delta, growth, decay, or before/after comparison & Two or more grounded measurements support the change \\
\taxsep
\taxrow{Interpolation within support} & Ask for an approximate value inside a visible interval or plotted range & Tolerance is specified\opsep extrapolation is rejected unless explicitly bounded \\
\taxrow{Symbolic-numeric translation} & Convert visual relations into equations, expressions, inequalities, or substitutions & Variables bind to visual/textual evidence and simplification agrees with the final answer \\
\taxsep
\taxrow{Combinatorial counting} & Ask for counts under conditions, arrangements, paths, subsets, or grouped categories & Count domain is visually finite and all exclusions are explicit \\
\taxsep
\taxrow{Counterfactual within image constraints} & Modify one visible condition and ask for the resulting answer while holding other visible conditions fixed & Counterfactual only changes stated variables\opsep no external causal assumptions required \\
\taxsep
\taxrow{Distractor generation} & Add options or false leads that require visual elimination rather than language priors & Each distractor has a known violated constraint \\
\taxsep
\taxrow{Evidence localization} & Ask for the supporting cell, span, region, point, object, or diagram component in addition to the answer & Region/cell/span metadata is present and checkable \\
\taxsep
\taxrow{Difficulty calibration} & Increase steps, add conditions, or compose operations only until the expected solver pass rate lies in the target range & Rollout pass rate remains between selected thresholds \\
\taxsep
\taxrow{Verifiability conversion} & Prefer exact, normalized, bounded, or programmatically checkable answer forms over vague open-ended forms & A deterministic or high-confidence verifier can be attached \\
\taxsep
\taxrow{Ambiguity reduction} & Rewrite open questions into constrained, unique-answer prompts with explicit scope and answer format & Multiple plausible interpretations are eliminated by schema and visual grounding \\
\taxsep
\taxrow{Adversarial visual dependence} & Add a text-only distractor or require a visible detail that language priors alone cannot determine & Text-only probe fails or has low confidence \\
\end{longtable}
\endgroup

\subsection{Answer-Type and Reward Contracts}

\begingroup
\footnotesize
\setlength{\tabcolsep}{3pt}
\renewcommand{\arraystretch}{1.25}
\setlength{\heavyrulewidth}{1.1pt}
\setlength{\lightrulewidth}{0.5pt}
\begin{longtable}{@{}>{\raggedright\arraybackslash}p{0.15\linewidth}
                    >{\raggedright\arraybackslash}p{0.43\linewidth}
                    >{\raggedright\arraybackslash}p{0.35\linewidth}@{}}
\caption{Answer-type-conditioned operators and reward contracts.}
\label{tab:app-answer-operators}\\
\toprule
\taxhead{Answer route} & \taxhead{Answer-conditioned evolution operators} & \taxhead{Verification / reward contract} \\
\midrule
\endfirsthead
\midrule
\taxhead{Answer route} & \taxhead{Answer-conditioned evolution operators} & \taxhead{Verification / reward contract} \\
\midrule
\endhead
\endfoot
\bottomrule
\endlastfoot
\taxrow{Multiple choice} & Preserve legal options, regenerate more discriminative distractors, or convert direct lookup into option-wise constraint satisfaction & Final answer must be a valid option\opsep verifier performs blind elimination over all options \\
\taxsep
\taxrow{Integer or decimal} & Force arithmetic over visible quantities\opsep add unit conversion\opsep require rounded, exact, or bounded numeric form & Normalize units, signs, separators, and tolerance\opsep programmatic check when possible \\
\taxsep
\taxrow{Fraction, ratio, percentage} & Ask for normalized ratio, proportion, relative frequency, percentage point difference, or percent change & Canonicalize equivalent forms\opsep verify numerator, denominator, and before/after quantities \\
\taxsep
\taxrow{Symbolic expression} & Ask for expression derivation, simplification, parameter relation, or symbolic substitution into visual constraints & Run symbolic simplification or numeric substitution\opsep bind each variable to evidence \\
\taxsep
\taxrow{Equation or inequality} & Ask for an equation, threshold, feasible interval, or inequality relation satisfying visible constraints & Check algebraically or by sampled substitution\opsep reject missing domain constraints \\
\taxsep
\taxrow{Coordinate, point, or interval} & Ask for an intercept, coordinate, range, interval, region boundary, or approximate plotted value & Normalize coordinate frame\opsep apply resolution-aware tolerance\opsep require axis evidence \\
\taxsep
\taxrow{Set, list, or ranking} & Ask for all objects satisfying a condition, ordered ranks, top-$k$, or grouped categories & Specify whether order matters\opsep normalize aliases\opsep verify completeness and unsupported-item absence \\
\taxsep
\taxrow{Boolean or verdict} & Ask whether a visual-mathematical statement is true, false, satisfiable, increasing, connected, or consistent & Require a cited counterexample or supporting constraint\opsep reject subjective yes/no prompts \\
\taxsep
\taxrow{Region, span, cell, or bounding evidence} & Ask for the relevant crop, text span, table cell, plotted point, graph node, or diagram component supporting the answer & Require coordinates, cell IDs, OCR spans, graph IDs, or crop references in metadata \\
\taxsep
\taxrow{Short categorical answer} & Convert open wording into a closed ontology such as color, shape, trend, state, relation type, component class, or direction & Normalize aliases\opsep reject subjective labels or categories not visually grounded \\
\taxrow{Free-form rationale} & Keep final answer short but require a concise evidence trace or step sketch & Use for SFT unless an extracted final answer has a deterministic checker \\
\taxsep
\taxrow{Program assertion} & Ask for an answer that can be represented as executable assertions over extracted quantities & Reward is assertion pass/fail with format-error penalty \\
\end{longtable}
\endgroup

\subsection{From Taxonomy to the 12-Topic Implementation}

The design-space taxonomy above admits a wide combinatorial cross product of (visual route) $\times$ (reasoning skill) $\times$ (answer type) $\times$ (verifier contract). Rather than instantiating this cross product at runtime, our current implementation discretizes the routing axes into 12 topic tags and pre-bundles, for each tag, the recommended reasoning operator, the recommended answer-type operator, and the matching verifier contract into a single evolution prompt template. Concretely, the 12 tags are: \texttt{OCR}, \texttt{Image Description}, \texttt{Detection}, \texttt{Analysis}, \texttt{Content Creation}, \texttt{Suggestions}, \texttt{Summarization}, \texttt{Logical Reasoning}, \texttt{Science-Related}, \texttt{Concept Extraction}, \texttt{Medical Imaging}, and \texttt{Scene Understanding}; their natural-language descriptions and example image families are documented in the released code. We chose this discretization for three reasons. First, a 12-topic vocabulary is small enough that each topic's evolution prompt can be authored and audited by hand, and each gate's failure modes can be enumerated; the full cross product would be impractical to maintain at this scale. Second, fixing the operator bundle per topic eliminates a class of compositional errors that arise when independently sampled operators produce internally inconsistent prompts (e.g., a multiple-choice answer-type operator attached to a problem whose answer is a continuous quantity). Third, adopting one prompt per topic makes the rejection reasons of the question-level gate easier to attribute back to the topic, which in turn supports per-topic threshold tuning. The trade-off is reduced expressivity: prompts that legitimately need to mix two routes (e.g., a document page combining a formula with a chart) are currently authored as part of the most fitting single topic rather than composed from independent route operators. Closing this gap with a verifier-aware operator-composition mechanism is left to future work.

\subsection{HTV-Agent Decider Prompt (verbatim)}
\label{app:decider-prompt}

The conflict-aware decider in Section~\ref{sec:agents} is realized as a constrained LLM call (no tool use, no code execution). The system prompt below is the verbatim template used during \rldata\ construction; the user prompt is filled in with the question, optional context, options (if any), the solver's hypothesized answer, and the verifier's report, in that order.

\noindent\textbf{System prompt (verbatim).}
\begin{lstlisting}[style=promptblock]
You are a final judge. You will see an initial answer from a Solver and a verification report from a Verifier. Your job is to decide the final answer.
Do not call tools, functions, notebooks, plugins, Python, or code interpreters.
Instructions:
1. Briefly summarize the verification's conclusion (1--2 sentences).
2. Assess the QUALITY of the verification: is it logically sound, or does it contain self-contradictions, arithmetic errors, or unsupported claims?
3. Decide the final answer:
   -- If verification is high-quality AND explicitly rejects the initial answer, trust the verification.
   -- If verification is low-quality or self-contradictory, trust the initial answer.
   -- If they agree, keep the answer.
4. If the Solver and Verifier disagree AND you are not confident in either, output <require_rethink>true</require_rethink>.
5. Output your exact final answer inside the tag
   <final_answer>X</final_answer>.
   Examples: <final_answer>A</final_answer> or <final_answer>42</final_answer>.
6. Output your confidence as <confidence>0--100</confidence>.
\end{lstlisting}

\noindent\textbf{User prompt template.}
\begin{lstlisting}[style=promptblock]
Question:
{question}
Context: {context}  (omitted if empty)
Options: {choices}  (omitted if not multiple choice)
Solver's answer: {hypothesis}
Verification report:
{verification_text}
\end{lstlisting}

\noindent\textbf{Post-processing.} The decider's response is parsed by regular expressions for \texttt{<final\_answer>}, \texttt{<confidence>}, and \texttt{<require\_rethink>}. If \texttt{<require\_rethink>true</require\_rethink>} is present, the sample is marked as unresolved and rejected by the deterministic gate rather than being forced into either \sftdata\ or \rldata. The decider is queried with sampling temperature $0$, so the decision is deterministic given the same solver/verifier inputs.

\section{Visual-Dependence Case Studies}
\label{app:cases}

\paragraph{Two illustrative flips.} We describe two side-by-side reasoning-trace case studies referenced in Section~\ref{sec:exp_error}. In the first, a geometry-diagram problem, RL-Origin asserts an unsupported angle while RL-Evol\,+\,Verifier extracts the correct measurement directly from the figure. In the second, a chart problem, RL-Origin reads the wrong column label while RL-Evol\,+\,Verifier identifies the column via its position relative to a labelled axis. Both pairs are drawn from the 12 paired flips identified in the 60-sample pilot (Layer~2, Section~\ref{sec:exp_error}).

\section{Prompt-Evolution Case Studies}
\label{app:evol_cases}

\begin{figure*}[t]
\centering

\begin{minipage}[c]{0.30\linewidth}
  \centering
  \includegraphics[width=\linewidth]{evol_case_imgs/case_moon.jpg}\\[2pt]
  {\scriptsize\textit{Lunar-phase diagram}}
\end{minipage}\hfill
\begin{minipage}[c]{0.66\linewidth}
  \begin{seedbox}
  \footnotesize\textsf{Q:} Identify the moon phase in which the whole side facing earth is lit. \quad\textsf{A:} \textbf{Full Moon}
  \end{seedbox}
  \vspace{2pt}
  \begin{evolbox}
  \footnotesize\textsf{Q:} Assuming the synodic period of the Moon is exactly $29.5$ days, calculate the total duration, in days, that the visible illumination is both strictly greater than the percentage shown during the \emph{First Quarter Half} phase and strictly less than the percentage shown during the \emph{Full Moon} phase.\\[2pt]
  {\scriptsize\textcolor{evolpost}{\textbf{Reasoning:}} First Quarter $=50\%$, Full Moon $=100\%$; the band $50\%\!<\!P\!<\!100\%$ spans exactly the Waxing- and Waning-Gibbous quarters, i.e.\ $T/2 = 29.5/2$.}\\[1pt]
  \textsf{A:} \ansok{14.75}
  \end{evolbox}
\end{minipage}

\vspace{6pt}

\begin{minipage}[c]{0.30\linewidth}
  \centering
  \includegraphics[width=0.82\linewidth]{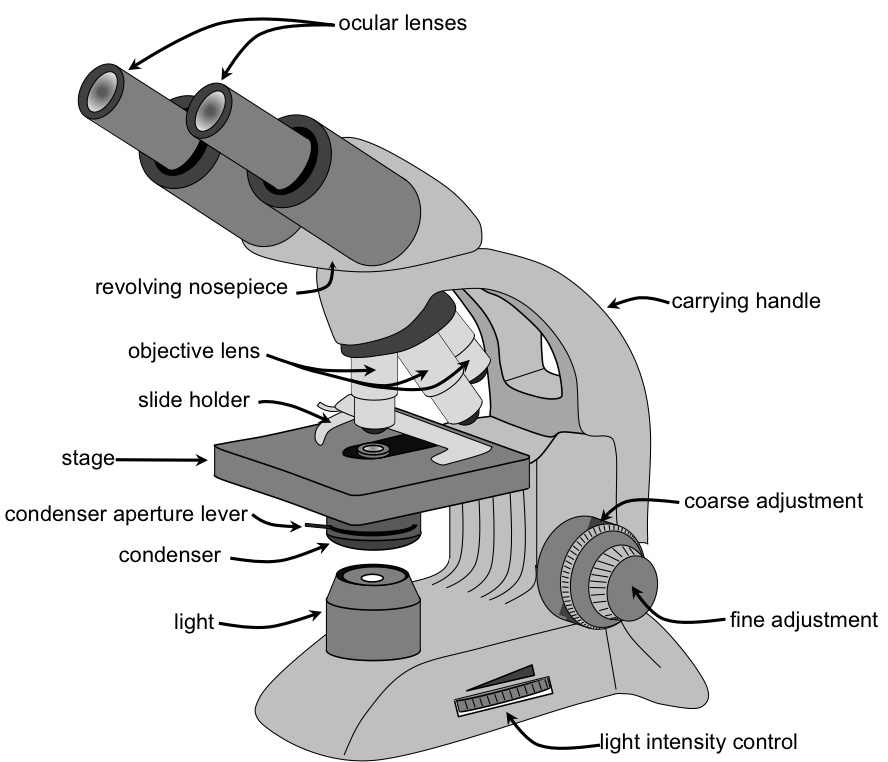}\\[2pt]
  {\scriptsize\textit{Labeled compound microscope}}
\end{minipage}\hfill
\begin{minipage}[c]{0.66\linewidth}
  \begin{seedbox}
  \footnotesize\textsf{Q:} Where would you look through this microscope? \quad\textsf{A:} \textbf{Ocular lenses}
  \end{seedbox}
  \vspace{2pt}
  \begin{evolbox}
  \footnotesize\textsf{Q:} Starting from a microscope that is powered off, unfocused, and set to the lowest-power objective, what is the minimum number of \emph{distinct, labeled mechanical controls} (knobs, levers, sliders) that must be manipulated to obtain a focused, properly illuminated view at the second-highest magnification?\\[2pt]
  {\scriptsize\textcolor{evolpost}{\textbf{Reasoning:}} distinct controls used $=$ \{light-intensity control, coarse adjustment, fine adjustment, condenser-aperture lever, revolving nosepiece\}; re-uses do not add new controls.}\\[1pt]
  \textsf{A:} \ansok{5}
  \end{evolbox}
\end{minipage}

\vspace{6pt}

\begin{minipage}[c]{0.30\linewidth}
  \centering
  \includegraphics[width=0.62\linewidth]{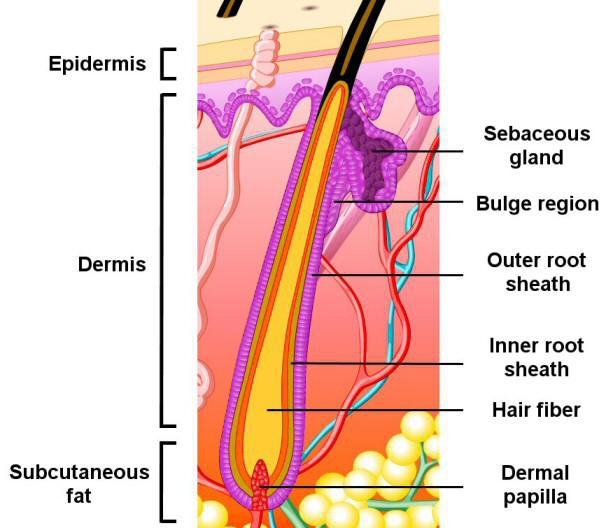}\\[2pt]
  {\scriptsize\textit{Skin / hair-follicle anatomy}}
\end{minipage}\hfill
\begin{minipage}[c]{0.66\linewidth}
  \begin{seedbox}
  \footnotesize\textsf{Q:} In which layer of the skin does the sebaceous gland lie? \quad\textsf{A:} \textbf{Dermis}
  \end{seedbox}
  \vspace{2pt}
  \begin{evolbox}
  \footnotesize\textsf{Q:} Compute the ratio of (A) the number of labeled hair-follicle components that physically extend past the Dermis/Subcutaneous-fat boundary, to (B) the number of labeled structures whose entirety is confined strictly within the Dermis.\\[2pt]
  {\scriptsize\textcolor{evolpost}{\textbf{Reasoning:}} $A=4$ (outer/inner root sheath, hair fiber, dermal papilla); $B=2$ (sebaceous gland, bulge region); ratio $=4/2$.}\\[1pt]
  \textsf{A:} \ansok{2}
  \end{evolbox}
\end{minipage}

\caption{\textbf{Type-aware prompt evolution on science-diagram seeds.} For each case, the grey box is the original single-step recognition seed and the purple box is the evolved, image-grounded multi-step reasoning prompt produced by the route-specific operators (\S\ref{sec:evolution}), annotated with the key steps of its \agent-verified solution. Evolved answers remain short and deterministically checkable, satisfying the reward-checker contract for \rldata.}
\label{fig:evol_cases}
\end{figure*}

To make the effect of type-aware prompt evolution (\S\ref{sec:evolution}) concrete, Figure~\ref{fig:evol_cases} shows three seed--evolved pairs drawn from science-diagram seeds. In each case the \emph{before} prompt is a single-step recognition or lookup query whose answer can often be retrieved from text priors or a single label, whereas the \emph{after} prompt produced by the route-specific operators requires multi-step reasoning that is grounded in spatial relations, labels, and quantities that must be read from the image. The evolved answer is short and deterministically checkable, so it admits the reward-checker contract required for \rldata\ (\S\ref{sec:training_signal}). We annotate each evolved prompt with the key reasoning steps from the \agent-verified solution.

\section{Iterative Construction Algorithm}
\label{app:algorithm}

Algorithm~\ref{alg:verievol} specifies the iterative construction procedure used by \method. It operates one seed sample at a time: after routing, candidate questions are generated under route-specific operators, filtered through the question-level gate $\phi_q$, screened by \agent, filtered again through the answer-level gate $g$, materialized into \sftdata, and (if the verifier supports deterministic checking and the rollout pass rate falls in the useful range) additionally materialized into \rldata.

\begin{algorithm}[h]
\caption{Iterative construction of the SFT and RL training corpora.}
\label{alg:verievol}
\begin{algorithmic}[1]
\Require seed pool $\mathcal{D}_0$, evolution operator library $\mathcal{O}$, question gate $\phi_q$, answer gate $g$, rollout budget $N$
\Ensure $\mathcal{D}_{\mathrm{SFT}}$, $\mathcal{D}_{\mathrm{RL}}$
\For{each seed sample $(I,q,a) \in \mathcal{D}_0$}
  \State infer routing tags $z=(t,y,u,c)$ using image, OCR, layout, and question features
  \State generate candidates $\mathcal{Q}=\{q'_j\}_{j=1}^{K}$ with operators in $\mathcal{O}_z$
  \For{each $q'_j \in \mathcal{Q}$}
    \If{$\phi_q(I,q,q'_j)=0$} \State discard or regenerate $q'_j$; \textbf{continue} \EndIf
    \State run \agent to obtain verified answer $\hat{a}_j$, accepted reasoning trace $r_j$, evidence $e_j$, verifier $v_j$, metadata $m_j$
    \If{$g(\hat{a}_j,m_j)=0$} \State discard $q'_j$; \textbf{continue} \EndIf
    \State materialize $(I,q'_j,r_j,\hat{a}_j,e_j)$ into \sftdata
    \If{$v_j$ supports deterministic or high-confidence checking}
      \State estimate rollout pass rate $\rho_j$ with $N$ student samples
      \If{$\rho_{\min}<\rho_j<\rho_{\max}$} \State add $(I,q'_j,\hat{a}_j,v_j,m_j)$ to \rldata \EndIf
    \EndIf
  \EndFor
\EndFor
\end{algorithmic}
\end{algorithm}


\end{document}